\documentclass[sigconf]{acmart}
\AtBeginDocument{%
  \providecommand\BibTeX{{%
    \normalfont B\kern-0.5em{\scshape i\kern-0.25em b}\kern-0.8em\TeX}}}

\copyrightyear{2026}
\acmYear{2026}
\acmConference[CIKM '26] {Proceedings of the 35th ACM International Conference on Information and Knowledge Management}{November 7--11, 2026}{Rome, Italy.}
\acmBooktitle{Proceedings of the 35th ACM International Conference on Information and Knowledge Management (CIKM '26), November 7--11, 2026, Rome, Italy}

\usepackage{amsfonts, amsmath, amsthm}
\usepackage[ruled, vlined, linesnumbered]{algorithm2e}
\usepackage{graphicx}
\usepackage{multirow}
\usepackage{nccmath}
\usepackage{subfig}
\usepackage{xcolor, xspace}
\usepackage{placeins}
\usepackage{pifont}
\usepackage{colortbl}
\usepackage{mathtools}

\newtheorem{assumption}{Assumption}

\newtheorem{proposition}{Proposition}

\newtheorem{lemma}{Lemma}

\newtheorem{theorem}{Theorem}

\newcommand{\eg}{\textit{e.g.}\xspace}
\newcommand{\ie}{\textit{i.e.},\xspace}
\newcommand{\etc}{etc.\xspace}

\newcommand\figref[1]{Fig.~\ref{#1}}
\newcommand\algref[1]{Algorithm~\ref{#1}}
\newcommand\tabref[1]{Table.~\ref{#1}}
\newcommand\secref[1]{Sec.~\ref{#1}}
\newcommand\equref[1]{Eq.~(\ref{#1})}

\newcommand\lemmaref[1]{Lemma ~\ref{#1}}
\newcommand\propref[1]{Proposition ~\ref{#1}}
\newcommand\asmref[1]{Assumption ~\ref{#1}}
\newcommand\appref[1]{Appendix~\ref{#1}}

\newcommand{\fakeparagraph}[1]{\noindent\textbf{#1.}}

\newcommand{\sysname}{X-FED\xspace}

\newcommand{\schname}{PFL-EE\xspace}

\ifodd 0

\newcommand{\zimu}[1]{{\color{brown}{#1}}}

\else

\newcommand{\zimu}[1]{#1}

\fi

\newcommand{\citeSplit}{\cite{arivazhagan2019federated, collins2021exploiting, oh2022fedbabu,cheng2026progressive,pan2026himac}\xspace}

\newcommand{\summ}{\sum_{j=1}^m}
\newcommand{\sumn}{\sum_{i=1}^n}

\begin{document}

\title{Federated Personalization of Early-Exit Networks}

\author{Boyi Liu}
\affiliation{%
  \institution{City University of Hong Kong Shenzhen Research Institute}
  \city{Shenzhen}
  \country{China}
}
\affiliation{%
  \institution{Department of Data Science, City University of Hong Kong}
  \city{Hong Kong}
  \country{China}
}
\affiliation{%
  \institution{SKLCCSE, Beihang University}
  \city{Beijing}
  \country{China}
}
\email{boy.liu@my.cityu.edu.hk}

\author{Zimu Zhou}
\affiliation{%
  \institution{City University of Hong Kong Shenzhen Research Institute}
  \city{Shenzhen}
  \country{China}
}
\affiliation{%
  \institution{Department of Data Science, City University of Hong Kong}
  \city{Hong Kong}
  \country{China}
}
\email{zimuzhou@cityu.edu.hk}

\author{Cheng Fang}
\affiliation{%
  \institution{Department of Data Science, City University of Hong Kong}
  \city{Hong Kong}
  \country{China}
}
\email{cheng.fang@my.cityu.edu.hk}

\author{Yongxin Tong}
\affiliation{
  \institution{SKLCCSE, Beihang University}
  \city{Beijing}
  \country{China}
}
\email{yxtong@buaa.edu.cn}

\begin{abstract}

Personalized Federated Learning (PFL) excels at tailoring client-specific models, which is particularly critical for decentralized and heterogeneous data environments, yet existing methods produce static models with a fixed tradeoff between accuracy and efficiency.
This inherent static nature limits their ability to adapt to inference demands that vary with context and resource availability, posing a challenge for real-world deployment.
Early-exit networks (EENs), which enable adaptive inference via intermediate classifiers, offer a promising solution.
However, integrating EENs into PFL introduces two intertwined conflicts: client-wise heterogeneity across clients and depth-wise interference arising from conflicting exit objectives.
Prior studies fail to resolve both conflicts simultaneously, leading to suboptimal performance.
In this paper, we propose \sysname, a novel Conflict-Aware Cross-Client Federated Exit Distillation framework that jointly addresses both client- and depth-wise conflicts while extending PFL to early-exit networks.
At its core, \sysname employs a progressive, depth-prioritized student coordination mechanism that mitigates interference among shallow and deep exits while enabling effective personalized knowledge transfer across clients.
Furthermore, we introduce a client-decoupled formulation that reduces communication overhead with theoretical soundness.
Extensive evaluations on various datasets show that compared to state-of-the-art PFL and PFL-EE methods, \sysname achieves higher accuracy while reducing inference costs by 30.79\%–46.86\%.


\end{abstract}



\maketitle

\begin{figure}[t]
    \centering
    \includegraphics[width=0.42\textwidth]{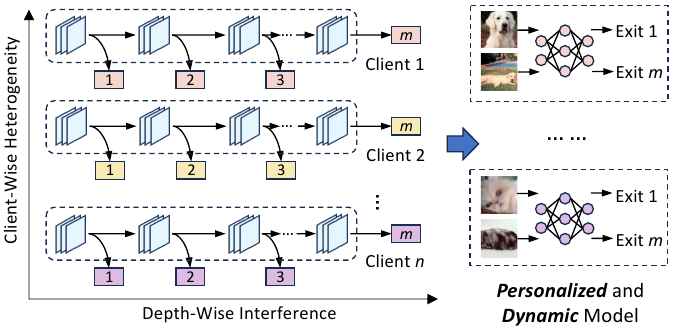}
  \caption{Motivation for personalized federated learning of early-exit networks (\schname).} 
  \label{fig:frame}
\end{figure}

\section{Introduction}
\label{sec:intro}

Personalized Federated Learning (PFL) \cite{tan2022towards} excels at leveraging the \textit{decentralized}, \textit{heterogeneous} data generated from IoT devices for intelligent applications such as smart healthcare \cite{ouyang2024admarker}, 
traffic prediction \cite{zhang2025fedmetro} and personalized assistants \cite{liang2026fedmosaic}. 
It enables collaborative model training across a network of IoT devices (clients) while tailoring models to each client's unique data distribution.
PFL often achieves higher accuracy than both generic federated learning (GFL) of a single global model \cite{yang2019federated,mcmahan2017communication,wang2023distribution,wei2025efficient} and local training on limited data, as it effectively integrates shared knowledge with client-specific insights.

Despite their improved accuracy, these personalized models are inherently \textit{static}, with a fixed tradeoff between accuracy and efficiency. 
IoT applications, however, often process continuous data streams in dynamic environments where inference requirements change with context, workload, or resource availability \cite{han2021dynamic}. 
For instance, an image classifier in autonomous driving must prioritize ultra-low-latency predictions to identify immediate hazards, such as nearby pedestrians or vehicles, while ensuring accurate recognition of distant objects or under complex lighting. 
Dynamic models, like early-exit networks (EENs) \cite{teerapittayanon2016branchynet, rahmath2024early}, address this challenge by attaching intermediate classifiers (\ie early exits) to the backbone, enabling inference to terminate at shallower exits when appropriate. 
This reduces latency while preserving accuracy for simpler inputs, making EENs ideal for IoT applications with fluctuating demands.

Can we effectively train both \textit{personalized} and \textit{dynamic} models in federated learning (see \figref{fig:frame})? 
Consider $n$ clients, each training personalized EENs with $m$ exits. 
This setup requires optimizing $mn$ objectives simultaneously, which introduces challenges beyond the \textit{scale}. 
Two types of exit \textit{conflicts} complicate the training process. 
The first is \textit{client-wise heterogeneity}, where variations in local data distributions necessitate careful decisions on what to share across clients versus what to personalize \cite{tan2022towards}. 
The second is \textit{depth-wise interference}, caused by conflicting backpropagation signals between shallow and deep exits during joint training \cite{laskaridis2021adaptive}.
These conflicts create intricate interdependence among exits, making personalized federated training of EENs (PFL-EE) distinct from both personalized federated learning of single-exit networks (PFL) \cite{arivazhagan2019federated, collins2021exploiting, oh2022fedbabu, huang2021personalized, ye2023personalized} and generic federated learning of EENs (GFL-EE) \cite{ilhan2023scalefl, kim2023depthfl, lee2024recurrent}.

\begin{figure}[t]
    \centering
    \includegraphics[width=0.40\textwidth]{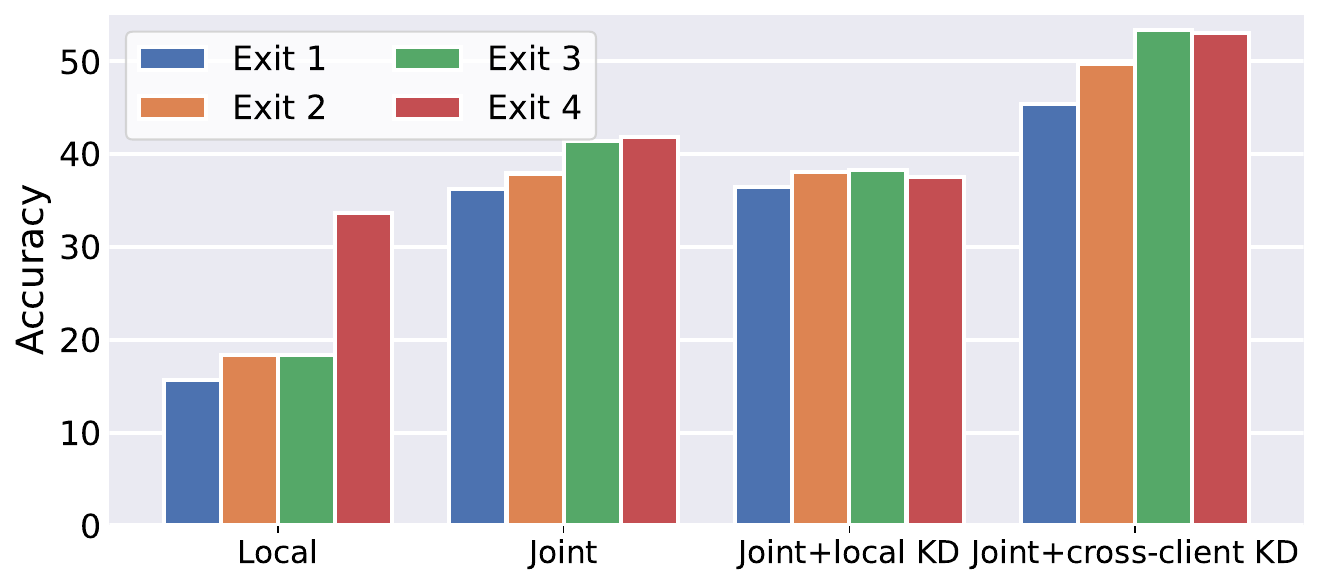}
  \caption{An empirical comparison of post-PFL exit training (local), direct extension of PFL to EENs (joint),  existing GFL-EE (joint + local KD), and our solution (joint + cross-client KD) to train personalized 4-exit EENs for 100 clients on CIFAR-100 under Dir(0.3).}
  \label{fig:empirical}
  \vspace{-2mm}
\end{figure}

Existing solutions fail to address both conflicts simultaneously. 
One naive approach is applying PFL to the last exit (and the backbone) \cite{arivazhagan2019federated}, followed by local training of shallower exits \cite{teerapittayanon2016branchynet}. 
However, this method underperforms because shallow exits also require knowledge across clients to compensate for the sparse local datasets (\textit{local} vs. \textit{joint} in \figref{fig:empirical}). 
Another option is to adopt recent federated training methods for global EENs \cite{ilhan2023scalefl}, which employ local knowledge distillation (KD) \cite{phuong2019distillation} to supervise the training of shallow exits. 
Yet, under severe data heterogeneity, which is intrinsic in PFL, local KD can degrade accuracy, performing worse than simple joint training (\textit{joint + local KD} vs. \textit{joint} in \figref{fig:empirical}).
Both strategies fail because they implicitly treat exits as analogous to clients, reducing PFL-EE to standard PFL with more clients, thereby overlooking the distinctions between client- and depth-induced exit conflicts.

In this paper, we present \sysname, a \underline{C}onflict-Aware 
\underline{C}ross-Client
\underline{F}ederated \underline{E}xit \underline{D}istillation framework for personalized federated learning of EENs. 
\sysname effectively resolves both client- and depth-wise conflicts, enabling shallow exits to distill personalized knowledge from the last exits across all clients.
At its core is a \textit{progressive}, \textit{depth-prioritized} coordination strategy for student exits. 
Specifically, \sysname gradually incorporates exits along the depth dimension rather than the client dimension, encouraging shallow exits to align with their corresponding teachers first, thereby reducing conflicts during subsequent distillation. 
The incremental inclusion also maintains manageable conflict levels throughout the training.
Following student selection, \sysname matches teacher exits to student exits based on inferred \textit{data similarity}, as is standard in PFL, to facilitate personalized knowledge transfer across clients. 
Furthermore, to avoid the communication overhead of cross-client KD, \sysname introduces a \textit{client-decoupled} formulation that achieves theoretical equivalence to cross-client KD while incurring a communication cost comparable to standard FedAvg \cite{mcmahan2017communication}.

Our main contributions are summarized as follows.
\begin{itemize}
    \item 
    To our knowledge, this is the first work on personalized federated learning of early-exit networks (PFL-EE).
    It advances PFL from static to dynamic models capable of high-fidelity and low-latency inference in evolving environments.
    \item 
    We propose \sysname, a unified distillation-based framework that addresses the dual challenges of client-wise heterogeneity and depth-wise interference in PFL-EE. 
    It harmonizes exit training through progressive, depth-prioritized student coordination and introduces a client-decoupled formulation to eliminate the communication overhead of direct cross-client distillation.
    \item 
    Evaluations on four datasets show that \sysname outperforms state-of-art PFL \cite{arivazhagan2019federated, collins2021exploiting, oh2022fedbabu, huang2021personalized, ye2023personalized, chen2022bridging, li2021ditto, xu2023personalized} and GFL-EE \cite{ilhan2023scalefl, kim2023depthfl} baselines in accuracy, while reducing the inference cost by up to $30.79$-$46.86\%$.
\end{itemize}

\section{Related Work}
\fakeparagraph{Personalized Federated Learning}
\zimu{PFL \cite{tan2022towards} trains personalized models rather than a single global model to handle non-IID data across clients.
Each model holds a client-specific classifier \citeSplit~or is a different sub-model of the global model \cite{zhang2023dm, chen2023efficient,ma2026harnessing} to retrain personalized parameters.}
For example, FedRep \cite{collins2021exploiting} divides the model into a global feature extractor and a local classifier.
Only the feature extractor is aggregated at the server.
pFedGate \cite{chen2023efficient} employs personalized masks to generalize different sub-models for clients.
Personalization can also be enforced via client clustering \cite{sattler2020clustered}, model regularization \cite{li2021ditto}, knowledge distillation \cite{zhang2021parameterized}, meta-learning \cite{fallah2020personalized}, 
\etc

\zimu{Existing PFL research trains personalized, static (single-exit) models, whereas we learn both personalized and dynamic models exemplified by early-exit networks.
For simplicity, we adopt the classifier-based strategy \citeSplit~for personalization.}

\fakeparagraph{Early-Exit Networks}
\zimu{EENs \cite{rahmath2024early} are dynamic neural networks \cite{han2021dynamic} that condition their computational depths on inputs.
They insert intermediate exits to the model, allowing early termination of inference on easy samples, thus improving inference efficiency.
An EEN can be optimized in model architectures \cite{teerapittayanon2016branchynet, huang2018multi, yang2020resolution}, exit policies \cite{kaya2019shallow, huang2024unlocking, regol2024jointly}, and training strategies \cite{phuong2019distillation, yu2023boosted, gong2024deep}. 
We employ the classic multi-branch architecture with a simple confidence-based exit policy \cite{teerapittayanon2016branchynet}, and mainly focus on EEN training.}

\zimu{The exits in EENs are often jointly trained to improve accuracy \cite{han2021dynamic, huang2018multi}, which can be further enhanced by knowledge distillation (KD) \cite{phuong2019distillation}.
Originally, EEN training assumes centralized settings.
Yet recent efforts \cite{kim2023depthfl, ilhan2023scalefl, lee2024recurrent, qu2025darkdistill} have explored federated learning of a global EEN with clients of heterogeneous resource constraints \cite{diaoheterofl}.
We investigate an orthogonal problem that trains personalized EENs with clients holding non-IID data.
Our solution is inspired by these studies \cite{kim2023depthfl, ilhan2023scalefl, lee2024recurrent} that utilize KD to harmonize exit training.
However, their KD is confronted within each local EEN. 
We show the necessity and propose a novel formulation of cross-client KD for federated learning of personalized EENs.}

\section{Problem Statement}
\label{sec:problem}

\fakeparagraph{Personalized Federated Learning (PFL)}
Assume $n$ clients $\{c_1$, $c_2$, $\ldots$, $c_n\}$ holding local datasets $\{D_1, D_2,\ldots, D_n\}$ \zimu{which are often not independent and identically distributed (non-IID)}.
PFL trains $n$ personalized, \zimu{\textit{single-exit}} models $\{ \theta_1,\ldots,\theta_n\}$ \zimu{in a standard federated setup \cite{mcmahan2017communication}}, by optimizing the following objective \cite{tan2022towards}:
\begin{equation}
    \min \sumn p_i F_i(\theta_i;D_i)
\end{equation}
where $p_i = \frac{|D_i|}{\sumn|D_i|}$, and $F_i$ is the local objective of client $i$.

\fakeparagraph{Early-Exit Networks (EENs)}
An EEN \cite{teerapittayanon2016branchynet} is a dynamic neural network that adjusts its computational depths at inference time based on inputs.
It consists of a backbone $\phi$ and $m$ exits $\{h_1,\ldots,h_m\}$.
The $m$ exits are often jointly trained on dataset $D$ to mitigate cross-exit interference and boost accuracy \cite{laskaridis2021adaptive, rahmath2024early}.
\begin{equation}
    \min \sum_{j=1}^m w_j F(\phi, h_j;D)
\end{equation}
where $\forall j$, the exit-wise weight $w_j = 1/m$, assuming each exit is equally important for inference \cite{huang2018multi, phuong2019distillation, ham2024neo, kim2023depthfl, lee2024recurrent}.


\fakeparagraph{Personalized Federated Learning of EENs (PFL-EE)}
\zimu{We train for each client $i$ a personalized, \textit{multi-exit} model $\theta_i = (\phi;h_{i1},\ldots,h_{im})$ in the federated manner, where $\phi$ is a backbone \textit{shared} among the $n$ clients, and $\{h_{i1},\ldots,h_{im}\}$ are $m$ personalized exits for client $i$, by optimizing the following objective:}
\begin{equation}
\label{equ:obj-eepfl}
    \min \sumn p_i \summ w_j F_i(\phi, h_{ij};D_i)
\end{equation}
It differs from prior studies as follows.
\begin{itemize}
    \item 
    \zimu{Traditional \textbf{PFL} \cite{arivazhagan2019federated, collins2021exploiting, oh2022fedbabu, huang2021personalized, ye2023personalized} optimizes $n$ \textit{client-wise} objectives, while we simultaneously optimize $mn$ objectives in both the \textit{client} and \textit{exit} dimensions.}
    \item 
    \zimu{EENs have been used to align features among local models of different depths in FL \cite{ilhan2023scalefl, kim2023depthfl, lee2024recurrent}, which we term as \textbf{GFL-EE}.
    It is primarily designed for IID data and learns a \textit{global} multi-exit model whereas we train $n$ \textit{personalized} multi-exit models.}
\end{itemize}

\fakeparagraph{Challenges}
\zimu{PFL-EE is more challenging than standard PFL and GFL-EE not only because of the increased \textit{scale of objectives} but also the need for a unified approach to handle two \textit{types of conflicts}.
\textit{(i)}
There is \textit{interference} among \textit{exits} due to the interplay of multiple backpropagation signals in joint exit training \cite{laskaridis2021adaptive}.
\textit{(ii)}
There is \textit{data heterogeneity} among \textit{clients} that demands careful balance between similarity and complementarity for effective federated training \cite{yan2024balancing}.}

\fakeparagraph{Scope}
We focus on the training of EENs.
We attach an exit per block to construct a classic multi-branch EEN architecture and apply a simple threshold-based exit policy \cite{teerapittayanon2016branchynet}.
For simplicity, FedAvg \cite{mcmahan2017communication} is used for model aggregation.


\section{Method}
\label{sec:method}

\newcommand{\vk}{\mathbf{k}}
\subsection{\sysname Overview}
\label{subsec:overview}
\sysname is a 
\underline{C}onflict-Aware
\underline{C}ross-Client 
\underline{F}ederated \underline{E}xit \underline{D}istillation framework for effective personalized federated learning of EENs.

\fakeparagraph{Principles}
\sysname is inspired by \cite{ilhan2023scalefl} that exploits the last exit as an anchor to harmonize the training of shallow exits.
This is achieved via knowledge distillation (KD) from the last exit to shallow ones \textit{within} the local EEN at each client.
We extend the idea to PFL-EE by allowing distillation from the last exits \textit{across} all clients to shallow exits \textit{across} all clients.

Specifically, let $\theta_{ij}=(\phi, h_{ij})$ be the sub-model that corresponds to backbone $\phi$ and exit $h_{ij}$ of client $i$ at \zimu{depth} $j$.
For a student model $\theta$, our \textit{cross-client} KD leverages all teacher models $\{\theta_{1m},\ldots,\theta_{nm}\}$ to improve $\theta$ via the following KD loss:
\begin{equation}
\label{equ:cc-kd}
    \mathcal{L}_{XKD}(\theta;x) = \sumn k_i D_{KL} \left(\theta_{im}(x) \| \theta(x) \right)
\end{equation}
where $D_{KL}$ is the KL divergence \cite{kullback1997information}, and $k_i$ is the distillation weight of teacher $i$ in \textit{cross-client} KD, satisfying $\sumn k_i =1$.
\begin{itemize}
    \item 
    \zimu{Cross-client KD is necessary for PFL-EE.
    Naively integrating local KD into PFL as GFL-EE \cite{ilhan2023scalefl, kim2023depthfl, lee2024recurrent} even yields a lower accuracy than PFL without KD (see \figref{fig:empirical}).}
    \item
    \zimu{Cross-client KD necessitates new student-teacher coordination mechanisms due to the increased scale and type of conflicts as explained in \secref{sec:problem}.}
\end{itemize}
\zimu{\sysname coordinates the distillation by \textit{prioritizing conflict management among exits over clients}.
\textit{(i)}
In the exit dimension, we fix the last exits from all clients as candidate teachers, and progressively include shallow exits as students, to mitigate the conflict exerted to the shared backbone.
\textit{(ii)}
In the client dimension, each student exit learns from the last exit beyond its own client, and it is matched with teacher exits from other clients based on their (estimated) data similarity for effective distillation.}

\begin{figure}[t]
    \centering
    \includegraphics[width=0.35\textwidth]{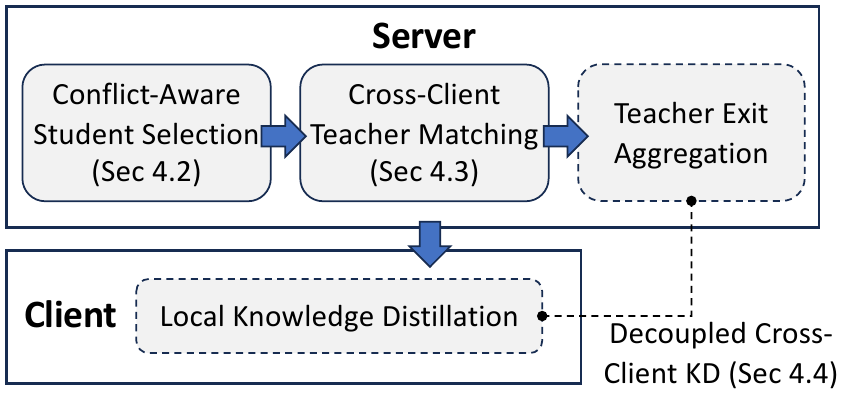}
  \caption{Overview of \sysname.}
  \label{fig:pipe}
  \vspace{-2mm}
\end{figure}

\fakeparagraph{Workflow}
\sysname integrates the above principles into a typical PFL pipeline (see \figref{fig:pipe}).
In each round, client $i$ trains the shared backbone $\theta$ and its personalized exits $\{h_{ij}\}$ on local dataset $D_i$ as:
\begin{equation}\label{equ:overall-obj}
    \min
    \sum_{j \in \mathbb{S}_i} w_j 
    \left( \mathcal{L}_{CE}(\theta_{ij};D_i)+ 
    \lambda \mathcal{L}_{XKD}(\theta_{ij};D_i)
    \right)
\end{equation}
where the first term is the standard cross-entropy loss, the second term is the cross-client KD loss defined in \equref{equ:cc-kd}, $\lambda$ is the hyperparameter controlling the proportion of KL loss, and $\mathbb{S}_i$ is the set of exits included for training at client $i$.

The key novelty of \sysname lies in the \textit{effective conflict management} when optimizing \equref{equ:overall-obj} via standard gradient descent.
\begin{itemize}
    \item 
    We progressively expand $\mathbb{S}_i$ for client $i$ via a \textit{conflict-aware student selection} scheme (\secref{subsec:scheduling}) to control exit-wise interference during joint exit training and distillation.
    \item 
    We adaptively set the distillation weights $\vk=[k_1,\ldots,k_n]$ based on the inferred similarities between student and teach exits via a \textit{cross-client teacher matching} strategy (\secref{subsec:teacher-matching}).
    \item 
    Naive implementation of the cross-client KD loss, \ie second term of \equref{equ:overall-obj}, needs exit parameters of all $n$ clients, which incurs excessive communication and violates privacy. 
    Hence, we propose an equivalent \textit{decoupled cross-client distillation} formulation (\secref{subsec:decoupled}) for practical deployment.
\end{itemize}

After local training, the server collects and aggregates the shared backbone from the clients as in other classifier-based PFL \citeSplit.
When the training converges, the shared backbone and the personalized exits are assembled as the final EEN for deployment to each client (see \figref{fig:frame}).

\subsection{Conflict-Aware Student Selection}
\label{subsec:scheduling}
\begin{figure}[t]
    \centering
    \includegraphics[width=0.45\textwidth]{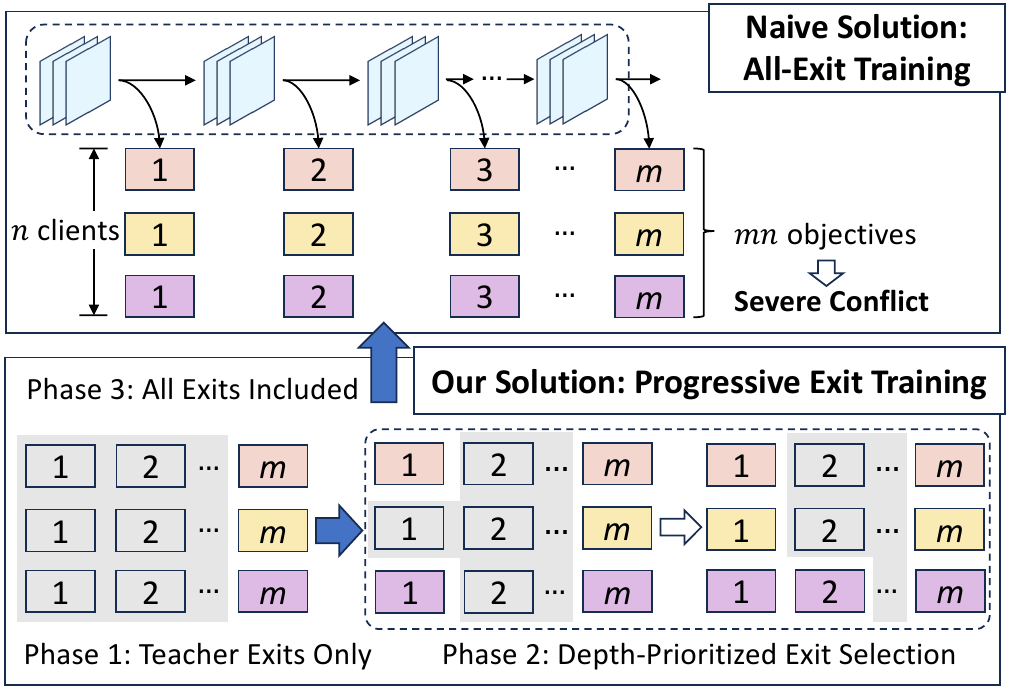}
  \caption{Conflict-aware student selection.}
  \label{fig:prog}
\end{figure}

\zimu{This module determines the set of exits $\mathbb{S}_i$ for local training and distillation at every client $i$ ($1\leq i\leq n$) in round $t$ ($1 \leq t \leq T$), with $T$ denoting the total number of rounds.
As mentioned in \secref{subsec:overview}, the last exits of all clients are employed as teachers.
Hence, $\mathbb{S}_i$ is initialized as $\{m\}$ for all client $i$ and round $t$.
We design a \textit{two-tier} selection scheme that gradually involves the remaining $(m-1)n$ exits as students for distillation (see \figref{fig:prog}).}

\zimu{Let the proportion of student exits in round $t$ be $R(t)$, which corresponds to $(m-1)nR(t)$ student exits.
We empirically set $R(t)=\min(\frac{2t}{T}, 1)$.
Other options are evaluated in \secref{subsubsec:impact-selection-rate}.
We gradually include more exits for training over rounds to avoid the substantial conflicts to optimize $mn$ exits all at once.
The concrete exit selection operates as follows.}

\fakeparagraph{Depth-Prioritized Exit Selection}
\zimu{In round $t$, we first include the $\lfloor (m-1)R(t)\rfloor$ \textit{shallowest} exits as students for every client $i$.
This step is essential for two reasons.
\begin{itemize}
    \item 
    If the $(m-1)n$ exits are treated as $(m-1)n$ clients rather than prioritize the depth, similarity-based client selection in PFL \cite{chen2024heterogeneity} tend to pick exits from a single client, since the data heterogeneity across clients is larger than that among exits of the same client.
    Yet this would cause the shared backbone to bias towards the selected client.
    \item
    Including shallower exits and aligning them with the teacher first will mitigate the conflicts in later distillation for deeper exits, since it encourages deeper exits to only focus on the knowledge not well-learned by shallow ones \cite{yu2023boosted}. 
\end{itemize}}

\fakeparagraph{Similarity-Aware Exit Selection}
For the remaining $K$=$(m-1)nR(t)-n\lfloor (m-1)R(t)\rfloor$ exits, we pick them from depth $\lfloor (m-1)R(t)\rfloor+1$ as standard similarity-based client selection in PFL.
We focus on client similarity because we have included the shallow exits of all exits, and thus sufficient client complementarity \cite{yan2024balancing}.
For two clients $i$ and $j$, we measure the similarity of their exits $l$ as $\delta_{ij}^l=\min(0, \cos(\theta_{il},\theta_{jl}))$.
The gradient-based proxy is common to measure data similarity among clients in FL \cite{sattler2020clustered, liu2024casa, liu2026towards}.

Then we iteratively remove the least similar exits from all exits $\mathcal{E}^l$ at depth $l$.
Specifically, we initialize the exit set $\mathcal{E} = \mathcal{E}^l$.
In each iteration, we greedily remove exit $e$ via the following criterion,
\begin{equation}
\label{equ:greedy-new}
    e = \arg \min _{e \in \mathcal{E}} \sum_{i \in \mathcal{E}} \delta^l_{ie}
\end{equation}
until the number of remaining exits $|\mathcal{E}|=K$.
We use the similarity of final exits $\cos(\theta_{im}, \theta_{jm})$ to approximate that of shallow exits $\cos(\theta_{il}, \theta_{jl})$ to reduce communication cost.

\subsection{Cross-Client Teacher Matching}
\label{subsec:teacher-matching}
\zimu{This module assigns distillation weights $\vk= [k_1,\ldots,k_n]$ to the $n$ teacher exits for the student exits $\{h_{ij}\}_{j=1}^{m-1}$ of client $i$.
As shown in \secref{sec:intro}, learning from other clients is critical to boost the accuracy of shallow exits, yet the teachers should be properly selected to avoid negative knowledge transfer due to inter-client data heterogeneity.
We maximize the knowledge sources by associating each student exit to all the $n$ teachers, and only distill matched knowledge via similarity-based teacher weighing.}
As mentioned in \secref{subsec:overview}, only the final exits in each EEN serve as teachers.

\fakeparagraph{Similarity-Based Teacher Weighting}
\zimu{We assign the weights $\vk= [k_1,\ldots,k_n]$ based on the inferred similarity in training data distributions between the student and teacher exits. 
A higher similarity results in a higher weight.
We apply the same gradient-based proxy as \secref{subsec:scheduling} and formulate the following objective:
\begin{equation}
\begin{aligned}
    \label{equ:solve-k}
    \arg \max_{\vk} \sumn k_i \cos(\theta_{im},\theta) - \mu \sumn \| k_i - \frac{1}{n} \|^2 \ \ s.t. \ \sumn k_i = 1
\end{aligned}
\end{equation}
where $\theta$ and $\theta_{im}$ are the student model and the teacher model, respectively.
The second term is a barrier regularization of $\vk$ controlled by hyperparameter $\mu$ \cite{zhang2021parameterized}.}

Note that \equref{equ:solve-k} approximately minimizes the transfer risk of KD, \ie student's risk after KD.
This is because the transfer risk bound for KD is related to the cosine angle of the teacher model and the student data space \cite{phuong2019towards}, which correlates to $\cos(\theta_{im}, \theta)$.

\fakeparagraph{Intra-Client Weight Approximation}
\zimu{\equref{equ:solve-k} demands solving $\vk$ for all student exits $\{h_{ij}\}_{j=1}^{m-1}$ of client $i$, which can be time-consuming.
Since the exit-wise heterogeneity is less notable than the client-wise heterogeneity in PFL \cite{chen2022pfl}, we enforce all student exits at client $i$ to share the same distillation weights $\vk$, which accelerates the preparation for distillation.}

\zimu{Specifically, we utilize the similarity between the teacher exit and the deepest student exit of client $i$ to approximate $\cos(\theta_{im},\theta)$.
Accordingly, we reduce \equref{equ:solve-k} from exit-wise to client-wise:}
\begin{equation}
\label{equ:quad-cvx}
\begin{aligned}
    \arg \max_\vk \ &\vk^\top c - \mu (\vk I-\frac{1}{n}I)^\top(\vk I-\frac{1}{n}I) \\
    &s.t. \ \ \textbf{1}^\top \vk=1; \vk\ge 0
\end{aligned}
\end{equation}
where $c=[\cos(h_{1m},h_{sm}),\ldots,\cos(h_{nm},h_{sm})]$ \zimu{for any student exit $s$ of client $i$}, and we solve \equref{equ:quad-cvx} via quadratic program.

\subsection{Decoupled Cross-Client Distillation}
\label{subsec:decoupled}
\begin{figure}[t]
    \centering
    \includegraphics[width=0.4\textwidth]{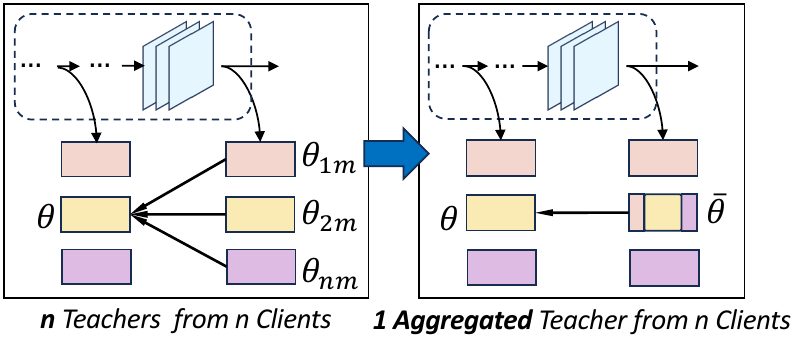}
  \caption{Decomposition of cross-client KD.}
  \label{fig:kd}
\end{figure}

\zimu{As mentioned in \secref{subsec:overview}, the cross-client KD formulated as \equref{equ:cc-kd} (second term in \equref{equ:overall-obj}) is impractical at the clients since it needs transmitting teacher exits of all other clients, which imposes high communication cost and violates privacy \cite{yang2019federated}.
One alternative is to shift the KD from clients to the server.
Yet it requires either a public dataset \cite{li2019fedmd, zhang2021parameterized} or a data generator \cite{zhu2021data} at the server, which is less favorable than client-side KD.
Instead, we propose an equivalent cross-client KD formulation without the need for other clients' raw model parameters at each client (see \figref{fig:kd}).}

\fakeparagraph{Client-Decoupled KD}
\zimu{We reformulate the cross-client KD in \equref{equ:cc-kd} as a two-step pipeline.
\begin{itemize}
    \item \textit{Teacher Exit Aggregation at Server}. 
    Since all clients share the same teacher exits (\ie the last exit of the $n$ clients) but differ in the distillation weights $\vk$, we can aggregate the $n$ teacher exits at the server:
    \begin{equation}\label{equ:decouple-aggr}
        \bar{h} = \sumn k_i h_{im}
    \end{equation}
    where the distillation weights $\vk$ are still calculated as \secref{subsec:teacher-matching}.
    \item \textit{Local KD at Clients}. 
    The aggregated teacher exit $\bar{h}$ and the shared backbone $\phi$ are sent to each client $i$ as the teacher model $\bar{\theta} = (\phi, \bar{h})$ for distillation with the student model $\theta$:
    \begin{equation}\label{equ:kd-all-obj}
        \mathcal{L}_{XKD}(\theta;x) =   D_{KL}(\bar{\theta}(x) \| \theta(x))
    \end{equation}
    Replacing the KD loss \equref{equ:overall-obj} by \equref{equ:kd-all-obj}, client $i$ performs local training and distillation as follows:
    
    \begin{equation}
    \label{equ:overall-obj-new}
    \mathcal{L}_i = 
    \sum_{j \in \mathbb{S}_i} w_j \mathcal{L}_{CE}(\theta_{ij};D_i) 
    + \lambda \sum_{j\in \mathbb{S}_i} D_{KL}(\bar{\theta}(D_i) \| \theta(D_i))
\end{equation}

    where we empirically set $\lambda=1$.
\end{itemize}}

\fakeparagraph{Equivalence to Cross-Client KD}
The two-step KD is equivalent to the objective in \equref{equ:cc-kd} via the following proposition.
\begin{proposition}[Equivalence of KD]\label{prop:equv}
Given an input sample $x$, the cross-client KD in \equref{equ:cc-kd} is equivalent to knowledge distillation from an aggregated teacher model $\bar{\theta} = (\phi, \bar{h})$ where $\bar{h} = \sumn k_i h_{im}$.
\begin{equation}
\begin{aligned}
    \theta^*
    = \arg \min_{\theta} \sumn k_i D_{KL}(\theta_{im}(x) \| \theta(x)) 
    = \arg \min_{\theta} D_{KL}(\bar{\theta}(x) \| \theta(x))
\end{aligned}
\end{equation}
\end{proposition}
That is, $\theta^*$ obtained through cross-client KD can be derived by minimizing the KL divergence between $\bar{\theta}_m$ and $\theta$.
The proof is deferred to \appref{proof:equv}.

\subsection{Putting It Together}
\label{subsec:together}

\algref{alg:overview} shows the workflow of \sysname.
In each round, the server randomly samples a client set $\mathbb{C}$ to participate in training (line 3).
For each sampled client, we further decide the student exits via conflict-aware student selection (line 4), and assemble a personalized teacher exit $\bar{h}_i$ for each student exit (line 5-7).
On receiving the shared backbone $\phi^t$ and the teacher exit $\bar{h}_i$, client $i$ updates its local model (line 10-15).
Then the backbone $\phi_i^t$ and final exit $h_{im}$ are uploaded to server for aggregation (line 16).
The aggregation of backbone aligns with FedAvg \cite{mcmahan2017communication} (line 17).
The final exits are kept at the server for future cross-client teacher matching.

\fakeparagraph{Convergence}
Under mild assumptions as \cite{li2020convergence}, \sysname converges to a data heterogeneity related constant after sufficient training round $T$ (details and proofs in \appref{proof:convergence}).


\begin{algorithm}[t]
    \KwIn{Clients' model parameters $\theta_1,...,\theta_n$} 
    \KwOut{Personalized models $\theta_1^t,...,\theta_n^t$}
    
    \For{round $t$}
    {
        \textcolor{blue}{\tt{// Server Execute}}
        
        sample clients set $\mathbb{C}$ for this round


        select exits $\{\mathbb{S}_i | i \in \mathbb{C}\}$ as \secref{subsec:scheduling}
        



    
    
        

        \For{client $i \in \mathbb{C}$}
        {
            
            solve $\vk$ via \equref{equ:quad-cvx}

            teacher exit $\bar{h}_{i} \gets$ aggregate via \equref{equ:decouple-aggr}
        }
        
        \textcolor{blue}{\tt{// Client Execute}}
        
        \For{client $i \in \mathbb{C}$}
        {
            receive $\phi_i^t,\bar{h}_i$ from server
            
            update local final exit $h_{im} \gets \bar{h}_i$
            

            update local backbone $\phi_i^t \gets \phi^t$
            

            calculate cross-client KD loss $\mathcal{L}_{XKD}$ via \equref{equ:kd-all-obj}
            
            calculate overall training objective $\mathcal{L}_i$ via \equref{equ:overall-obj-new}
            
            local training $\theta_i^{t+1} \gets \theta_i^t -\eta \nabla \mathcal{L}_i(\theta_i^t;D_i)$

            upload $\phi_i^t,h_{im}$ to server
        }

        aggregate backbone $\phi^{t+1} \gets \sum_{i \in \mathbb{C}} p_i \phi_i^{t}$
    }
    \caption{\sysname workflow}
    \label{alg:overview}
\end{algorithm}

\fakeparagraph{Communication Cost}
In \sysname, only the backbone and the last exit are transferred between the server and clients.
All shallow exits are kept locally.
The data exchanged per round is the same as FedAvg \cite{mcmahan2017communication}, \ie transmission of the full model.
Compared with classifier-based PFL \citeSplit, \sysname transfers slightly more data, \ie the teacher exits.
Yet the extra communication overhead is negligible.
For example, our experiments show that the communication cost of \sysname is merely $0.45\%$ higher than classifier-based PFL \citeSplit with ResNet-18 \cite{he2016deep}.

\fakeparagraph{Computational Cost}
\sysname adds extra server-side computation but matches GFL-EE baselines (\eg ScaleFL \cite{ilhan2023scalefl}) in client-side overhead, for they share the same number of local KD objectives.
The server-side computational cost of \sysname mainly comes from 
(i) conflict-aware student selection with $O(s^2n^2)$ (see \secref{subsec:scheduling}), 
(ii) cross-client teacher matching with $O(s^3n^3)$ (see \secref{subsec:teacher-matching}), and
(iii) teacher exit aggregation with $O(sn)$ (see \secref{subsec:decoupled}),
where $s$ is the sampling rate, and $n$ is number of clients.
Teacher matching with cubic level complexity dominates among the three. 
However, since it operates only over \textit{sampled clients} ($sn \ll n$), the actual cost is small. 
\secref{subsubseec:comp-teacher-matching} further provides an empirical study on the computational latency of teacher matching.

\section{Experiments}

\begin{table*}[t]
\caption{Overall performance of federated training of EENs. 
\textit{Averaged accuracy} of all exits are measured.
}
\label{tab:eefl}
\footnotesize
\small
\begin{tabular}{@{}llllllllll@{}}
\toprule
\multirow{2}{*}{Type} & \multirow{2}{*}{Method} & \multicolumn{2}{l}{CIFAR-10} & \multicolumn{2}{l}{CIFAR-100} & \multicolumn{2}{l}{TinyImageNet} & \multicolumn{1}{l}{AgNews} \\ 
\cmidrule(l){3-10} 
& & Dir(0.3) & Dir(0.1) & Dir(0.3) & Dir(0.1) & Dir(0.3) & Dir(0.1) & Dir(0.3) \\ 
\midrule
N/A                   
& Local-EE 
& $68.43_{\pm 12.52}$
& $81.92_{\pm 13.08}$
& $28.55_{\pm 5.76}$  
& $48.79_{\pm 9.74}$        
& $22.96_{\pm 12.91}$
& $38.28_{\pm 14.69}$    
& $71.91_{\pm 21.08}$
\\ 
\midrule
\multirow{4}{*}{GFL-EE}  
& FedAvg-EE  \cite{mcmahan2017communication}
& $49.46_{\pm 14.85}$
& $42.20_{\pm 17.11}$
& $31.53_{\pm 5.56}$
& $16.98_{\pm 5.94}$
& $19.55_{\pm 4.53}$
& $11.58_{\pm 5.96}$
& $25.11_{\pm 33.16}$
\\ 
& FedProx-EE  \cite{li2020federated}
& $62.05_{\pm 6.78}$ 
& $43.71_{\pm 19.43}$
& $30.88_{\pm 6.18}$
& $16.42_{\pm 5.61}$
& $16.80_{\pm 6.87}$
& $12.83_{\pm 4.82}$
& $27.67_{\pm 32.94}$
\\ 
& ScaleFL \cite{ilhan2023scalefl}
& $38.69_{\pm 7.71}$
& $26.75_{\pm 10.54}$
& $23.13_{\pm 4.56}$
& $9.83_{\pm 4.54}$
& $11.19_{\pm 3.28}$
& $7.87_{\pm 3.58}$
& $23.69_{\pm 31.96}$
\\
& DepthFL \cite{kim2023depthfl}
& $10.23_{\pm 17.99}$
& $9.87_{\pm 23.71}$
& $20.44_{\pm 7.52}$
& $12.56_{\pm 5.72}$
& $0.90_{\pm 1.49}$
& $0.55_{\pm 0.71}$
& $24.75_{\pm 34.06}$
\\ 
\midrule
\multirow{8}{*}{PFL-EE}
& FedPer-EE  \cite{arivazhagan2019federated}
& $71.39_{\pm 11.42}$
& $82.58_{\pm 11.77}$
& $39.39_{\pm 5.66}$
& $59.06_{\pm 8.20}$
& $27.55_{\pm 11.92}$
& $43.78_{\pm 13.24}$
& $77.18_{\pm 15.56}$
\\ 
& FedRep-EE  \cite{collins2021exploiting}
& $70.56_{\pm 11.49}$
& $82.71_{\pm 11.38}$
& $36.68_{\pm 5.77}$
& $59.26_{\pm 8.37}$
& $26.58_{\pm 11.96}$
& $44.03_{\pm 13.33}$
& $73.92_{\pm 19.50}$
&\\ 
& FedBABU-EE  \cite{oh2022fedbabu}
& $64.21_{\pm 13.51}$
& $75.23_{\pm 18.46}$
& $34.71_{\pm 9.07}$
& $53.45_{\pm 11.36}$
& $23.97_{\pm 13.70}$
& $37.63_{\pm 16.38}$
& $70.86_{\pm 22.17}$
\\ 
& FedAMP-EE   \cite{huang2021personalized}
& $47.25_{\pm 20.09}$
& $64.75_{\pm 25.76}$
& $24.98_{\pm 4.99}$
& $14.24_{\pm 9.50}$
& $16.04_{\pm 15.54}$
& $26.05_{\pm 17.21}$
& $62.08_{\pm 30.13}$
\\ 
& pFedGraph-EE  \cite{ye2023personalized}
& $50.19_{\pm 18.57}$
& $67.66_{\pm 24.10}$
& $10.52_{\pm 5.20}$
& $22.07_{\pm 8.75}$
& $9.20_{\pm 17.92}$
& $13.73_{\pm 20.91}$
& $71.27_{\pm 22.04}$
& \\ 
& FedRoD-EE   \cite{chen2022bridging}
& $66.36_{\pm 12.37}$
& $81.51_{\pm 13.35}$
& $42.59_{\pm 6.63}$
& $62.19_{\pm 8.35}$
& $28.91_{\pm 12.09}$
& $45.09_{\pm 12.85}$
& $77.24_{\pm 15.64}$
\\ 
& Ditto-EE   \cite{li2021ditto}
& $69.79_{\pm 12.66}$
& $82.60_{\pm 13.15}$
& $29.12_{\pm 5.71}$
& $49.19_{\pm 9.56}$
& $24.08_{\pm 12.68}$
& $40.23_{\pm 14.76}$
& $72.83_{\pm 20.10}$
\\ 
& FedPAC-EE  \cite{xu2023personalized}
& $71.60_{\pm 10.80}$
& $82.73_{\pm 13.70}$
& $47.45_{\pm 5.65}$
& $61.92_{\pm 7.89}$
& $33.96_{\pm 11.21}$
& $46.47_{\pm 12.65}$
& $76.17_{\pm 17.70}$
\\ 
\midrule
Ours
& \sysname 
& \textbf{72.67}$_{\pm 10.71}$
& \textbf{83.68}$_{\pm 13.55}$
& \textbf{50.41}$_{\pm 5.47}$
& \textbf{62.83}$_{\pm 8.64}$
& \textbf{36.27}$_{\pm 10.86}$
& \textbf{48.22}$_{\pm 12.59}$
& \textbf{78.69}$_{\pm 14.63}$
\\ 
\bottomrule
\end{tabular}
\end{table*}

\subsection{Experimental Setup}

\fakeparagraph{Baselines}
We mainly compare \sysname with representative GFL and PFL methods, including 
Local training, 
FedAvg \cite{mcmahan2017communication},
FedProx \cite{li2020federated},
FedPer \cite{arivazhagan2019federated},
FedRep \cite{collins2021exploiting},
FedBABU \cite{oh2022fedbabu},
FedAMP \cite{huang2021personalized},
pFedGraph \cite{ye2023personalized},
FedRoD \cite{chen2022bridging},
Ditto \cite{li2021ditto}, and
FedPAC \cite{xu2023personalized}.
For a fair comparison, we extend them to EENs.
\zimu{We also include GFL-EE methods dedicated to train a global EEN as baselines, including ScaleFL \cite{ilhan2023scalefl} and DepthFL \cite{kim2023depthfl}}.

\fakeparagraph{Datasets and Models}
We experiment with CIFAR-10/CIFAR-100 \cite{krizhevsky2009learning}, 
TinyImageNet \cite{chrabaszcz2017downsampled}, and 
AgNews \cite{zhang2015character}.
The data distribution adheres to a Dirichlet distribution, quantified by a hyperparameter $\alpha$.
We set $\alpha=0.3$ and $0.1$ following \cite{zhang2023dm}.
For CIFAR-10, we apply a 3-layer ConvNet.
For CIFAR-100 and TinyImageNet, we apply ResNet18.
For AgNews, we apply a 4-block Transformer.

\fakeparagraph{Configurations}
We simulate a federation of $100$ clients, with a sampling rate of $0.1$.
The communication round is set to $T=300$ for CIFAR-10, CIFAR-100, TinyImageNet, and $T=100$ for AgNews.
The local epoch is set to $E=5$ for CIFAR-10, CIFAR-100, TinyImageNet and $E=1$ for AgNews.
The batch size is fixed to $64$.
The learning rate is set to $0.1$, with a decay of $0.99$ per round.

\fakeparagraph{Metrics}
We assess the methods with three performance metrics:
\textit{(i)} \textit{Accuracy}: The model accuracy on local datasets measured at the terminated exit during inference. 
\textit{(ii)} \textit{Averaged Accuracy}: The average accuracy of all exits.
\textit{(iii)} \textit{Inference \zimu{Efficiency}}: The average number of \zimu{Multiply-Accumulate (MAC)} operations per input sample.

\subsection{Main Results}
\label{subsubsec:comparison-eefl}

\fakeparagraph{Averaged Accuracy}
\tabref{tab:eefl} reports the averaged accuracy of all exits across seven non-IID settings on four datasets.
The GFL-EE methods with local KD (\eg DepthFL and ScaleFL) do not perform well under severe non-IID settings.
This validates the observations in \figref{fig:empirical}.
Most PFL-EE methods outperform GFL-EE schemes, except FedAMP-EE and pFedGraph-EE.
\zimu{This is because these two baselines adopt personalized aggregation, which becomes ineffective with a low client sampling rate.}
\sysname improves the averaged accuracy over other PFL-EE methods by up to $9.57\%$, $21.29\%$, $12.19\%$ and $16.48\%$ for each dataset.

\begin{figure}[t]
  \centering
  \subfloat[CIFAR-10]{
    \includegraphics[width=0.23\textwidth]{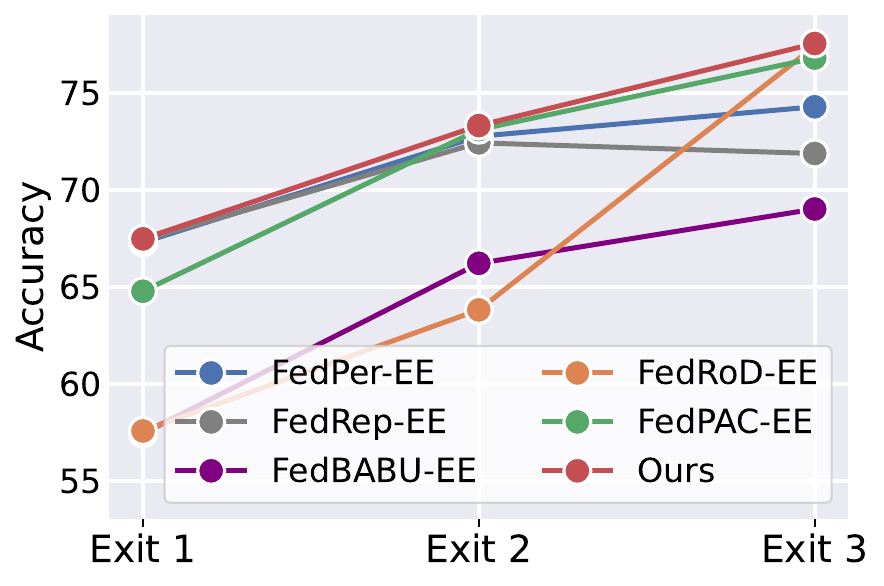}
    \label{fig:detail-cifar10}
  }
  \subfloat[CIFAR-100]{
    \includegraphics[width=0.23\textwidth]{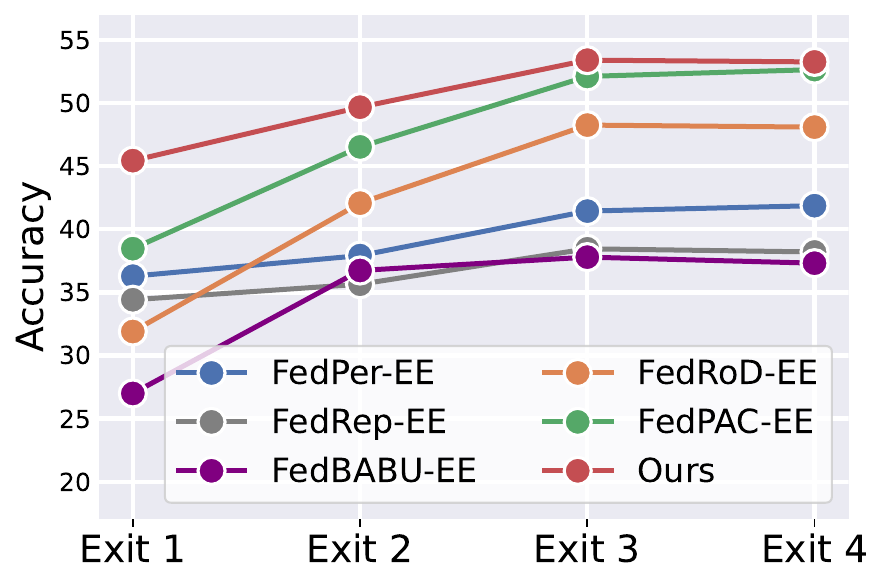}
    \label{fig:detail-cifar100}
  }
  \\
  \subfloat[TinyImageNet]{
    \includegraphics[width=0.23\textwidth]{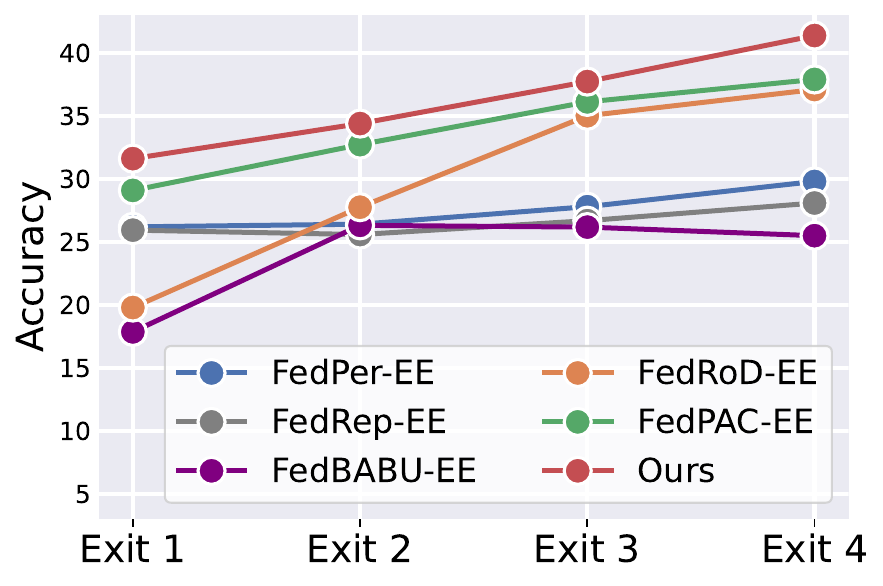}
    \label{fig:detail-tinyimagenet}
  }
  \subfloat[AgNews]{
    \includegraphics[width=0.23\textwidth]{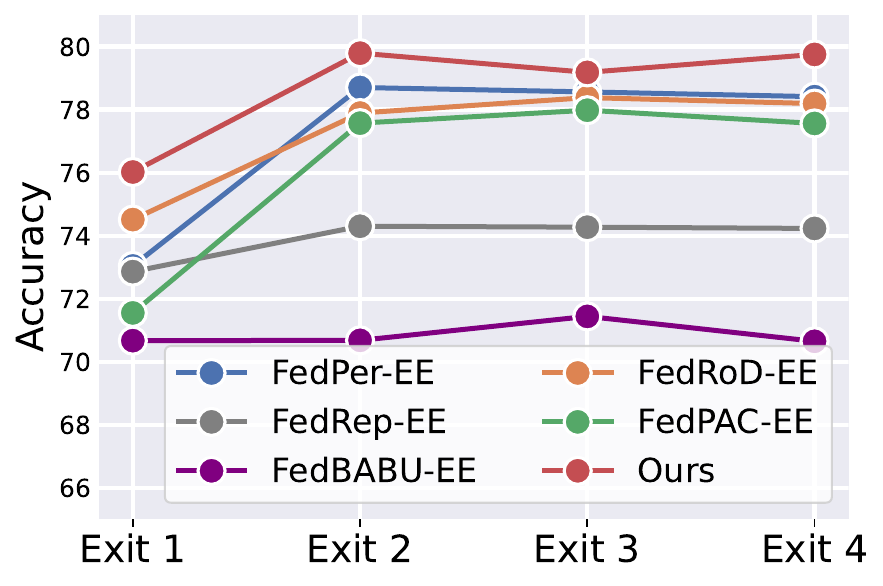}
    \label{fig:detail-agnews}
  }
  \caption{Per-exit accuracy of \sysname and five best-performing baselines from \tabref{tab:eefl}.}
  \label{fig:detail-all-exits}
\end{figure}

\fakeparagraph{Exit-Wise Accuracy}
\zimu{\figref{fig:detail-all-exits} zooms into the per-exit accuracy of the top five performing baselines in \tabref{tab:eefl}, \ie FedPer-EE, FedRep-EE, FedBABU-EE, FedRoD-EE, FedPAC-EE.}
We show the results under Dir(0.3).
\sysname outperforms the baselines \textit{in all exits}.
For example, on CIFAR-100, \sysname yields accuracy gains of 
$6.98$-$18.46\%$, 
$3.15$-$14.04\%$, 
$1.29$-$18.46\%$ and $0.61$-$16.11\%$ in each exit.


\begin{figure}[t]
  \centering
  \subfloat[CIFAR-10]{
    \includegraphics[width=0.23\textwidth]{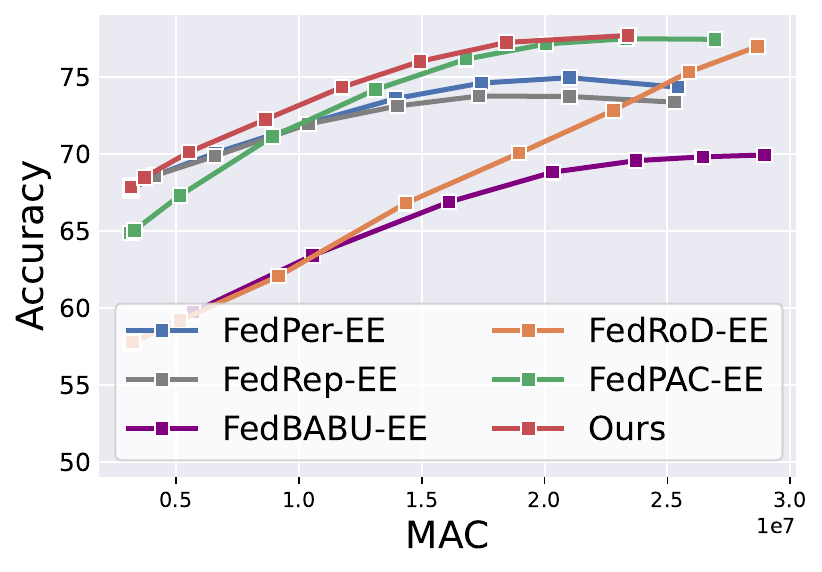}
    \label{fig:budget-cifar10}
  }
  \subfloat[CIFAR-100]{
    \includegraphics[width=0.23\textwidth]{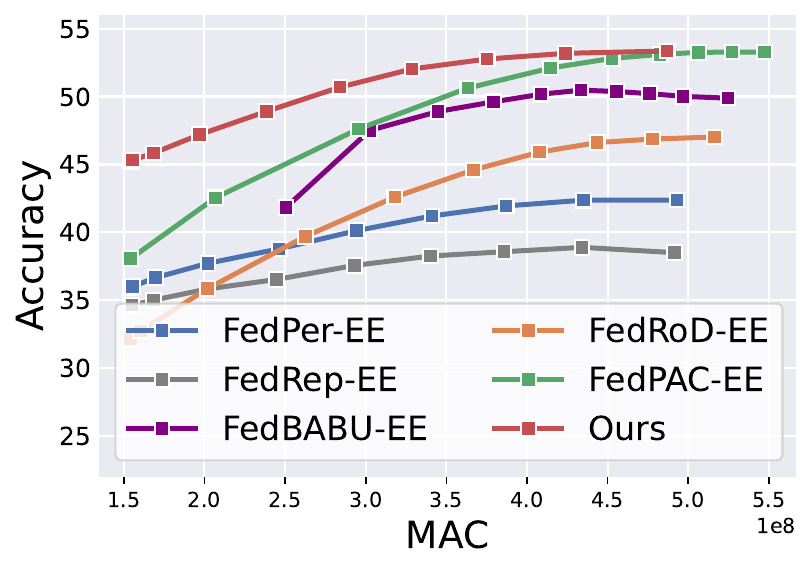}
    \label{fig:budget-cifar100}
  }
  \\
  \subfloat[TinyImageNet]{
    \includegraphics[width=0.23\textwidth]{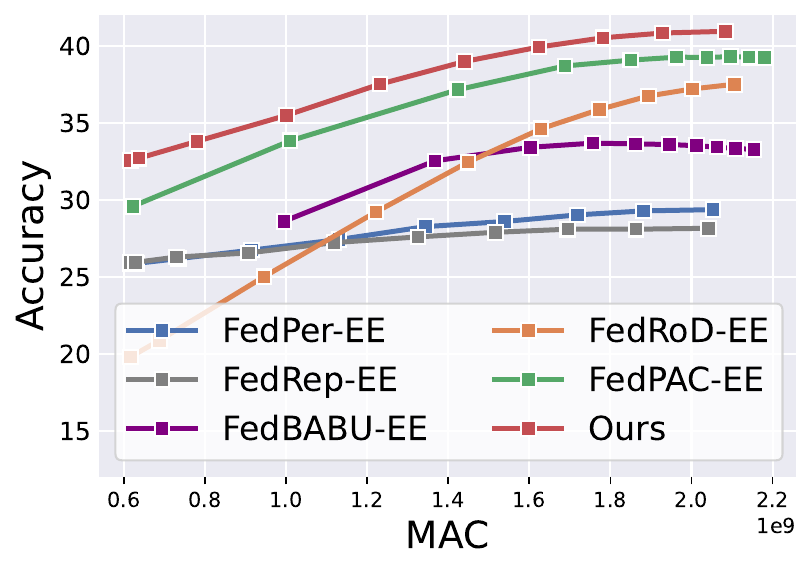}
    \label{fig:budget-tinyimagenet}
  }
  \subfloat[AgNews]{
    \includegraphics[width=0.23\textwidth]{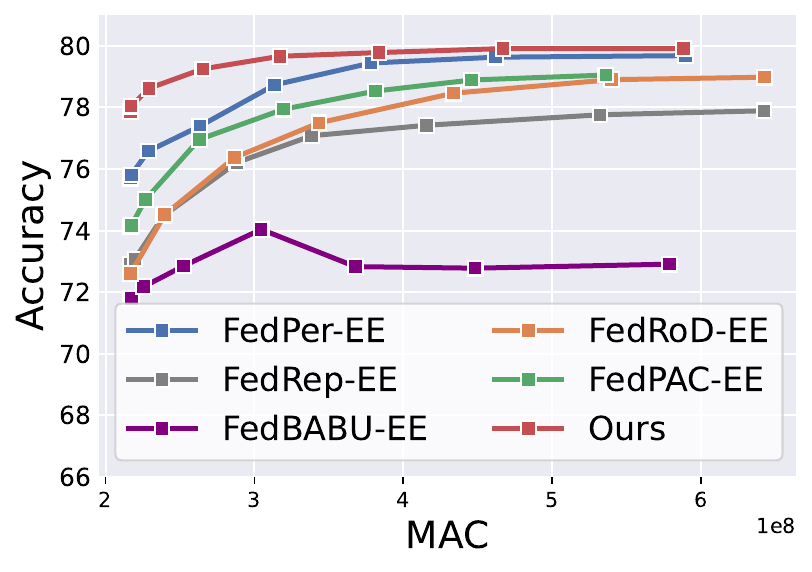}
    \label{fig:budget-agnews}
  }
  \caption{Accuracy-efficiency tradeoff of \sysname and five best-performing baselines from \tabref{tab:eefl}.}
  \label{fig:budget}
\end{figure}

\fakeparagraph{Accuracy-Efficiency Tradeoff}
We apply a naive exit policy \cite{huang2018multi} where the model terminates when the output confidence, \ie the maximum of the softmax output of an exit, exceeds a threshold $\epsilon$.
We compare \sysname with the same five baselines under the exit policy with different thresholds $\epsilon$ \zimu{to understand how they tradeoff between inference accuracy and efficiency.}

\figref{fig:budget} shows the measured accuracy and MAC.
On all four datasets, \sysname achieves a \textit{higher accuracy} at \textit{fewer MACs}.
Specifically, on CIFAR-10, for an accuracy of $70\%$, \sysname saves $0.2$-$5.5\times$ MACs.
On AgNews, for an accuracy of $78\%$, \sysname saves $0.2$-$2.8\times$ MACs.
On CIFAR-100, given a budget of $3.5\times10^8$ MACs, the accuracy of \sysname is $2.5$-$14.5\%$ higher.
On TinyImageNet, with $1.4\times10^9$ MACs, the accuracy of \sysname is $2$-$11\%$ higher.

\subsection{Micro-Benchmarks}
\label{subsec:micro}

\subsubsection{Benefit of EENs}
\zimu{Complementary to the main results, we directly compare \sysname with FL baselines that train a single-exit model to highlight the benefits of EENs.}
We exclude DepthFL and ScaleFL in this experiment since they are designed for EENs.
For \sysname, we set $\epsilon=0.8$ for CIFAR-10 and AgNews, and $\epsilon=0.6$ for CIFAR-100 and TinyImageNet.
We use different thresholds $\epsilon$ since datasets with fewer classes tend to reach higher confidence levels more easily.

\tabref{tab:pfl} shows the accuracy 
\sysname improves the accuracy over the PFL baselines by
$0.11$-$35.30\%$,
$4.88$-$40.61\%$, 
$0.99$-$28.92\%$ and 
$6.73$-$16.11$\%.
Due to the adaptive inference of EENs, the accuracy gains come with fewer MACs.
The averaged MAC per sample of the PFL baselines is $30.39M$, $557.94M$, $2231.64M$ and $811.57M$ on the four tasks, because they produce the same static model.
For \sysname, its averaged MAC per sample is $16.15M$, $300.45M$, $1545.69M$, $446.58M$, which reduces the inference cost by $46.15\%$, $44.73\%$, $30.74\%$ and $48.49\%$ for each dataset.

\begin{table}[t]
\caption{Accuracy of PFL methods.}
\label{tab:pfl}
\footnotesize
\begin{tabular}{@{}llllll@{}}
\toprule
Type & Method & CIFAR10 & CIFAR100 & TinyImageNet & AgNews \\ 
\midrule
N/A                   
& Local
& $68.67_{\pm 13.06}$
& $22.82_{\pm 5.37}$ 
& $22.07_{\pm 13.35}$
& $72.49_{\pm 20.85}$
\\ 
\midrule
\multirow{2}{*}{GFL}  
& FedAvg \cite{mcmahan2017communication}
& $63.20_{\pm 9.66}$
& $31.60_{\pm 4.94}$
& $19.21_{\pm 5.80}$
& $27.66_{\pm 32.95}$
\\ 
& FedProx \cite{li2020federated}
& $66.21_{\pm 6.98}$
& $30.80_{\pm 4.83}$
& $19.90_{\pm 5.41}$
& $26.69_{\pm 31.26}$
\\ 
\midrule
\multirow{8}{*}{PFL}
& FedPer \cite{arivazhagan2019federated}
& $73.98_{\pm 10.82}$
& $35.14_{\pm 6.10}$
& $26.87_{\pm 11.90}$
& $72.55_{\pm 20.66}$
\\ 
& FedRep  \cite{collins2021exploiting}
& $74.53_{\pm 10.70}$
& $30.62_{\pm 5.94}$
& $22.72_{\pm 12.73}$
& $71.79_{\pm 21.58}$
\\ 
& FedBABU  \cite{oh2022fedbabu}
& $75.20_{\pm 11.32}$
& $34.14_{\pm 6.72}$
& $28.14_{\pm 13.27}$
& $67.44_{\pm 25.78}$
\\ 
& FedAMP  \cite{huang2021personalized}
& $41.31_{\pm 22.16}$
& $26.08_{\pm 5.79}$
& $20.61_{\pm 13.35}$
& $63.17_{\pm 28.90}$
\\ 
& pFedGraph  \cite{ye2023personalized}
& $44.39_{\pm 22.19}$
& $10.60_{\pm 5.21}$
& $10.71_{\pm 17.50}$
& $63.28_{\pm 33.51}$
\\ 
& FedRoD  \cite{chen2022bridging}
& $75.33_{\pm 11.50}$
& $35.90_{\pm 6.67}$ 
& $27.18_{\pm 12.85}$
& $71.79_{\pm 21.58}$
\\ 
& Ditto  \cite{li2021ditto}
& $69.43_{\pm 13.11}$
& $23.96_{\pm 5.59}$
& $19.91_{\pm 13.37}$
& $71.03_{\pm 22.27}$
\\ 
& FedPAC  \cite{xu2023personalized}
& $76.50_{\pm 9.51}$
& $46.33_{\pm 6.25}$
& $38.64_{\pm 10.83}$
& $72.14_{\pm 21.25}$
\\ 
\midrule
Ours
& \sysname 
& \textbf{76.61}$_{\pm 9.16}$ 
& \textbf{51.21}$_{\pm 5.61}$ 
& \textbf{39.63}$_{\pm 10.33}$ 
& \textbf{79.28}$_{\pm 14.95}$ 
\\ 
\bottomrule
\end{tabular}
\end{table}

\subsubsection{Necessity of Cross-Client KD}
\label{subsubsec:necessity}
\zimu{Following the motivating experiment in \secref{sec:intro}, we provide a more detailed counterexample to justify the need for cross-client KD in PFL of EENs.
Specifically, we perform local KD on the PFL-EE baselines.}

\begin{table}[t]
\caption{Impact of local KD on PFL-EE baselines.}
\label{tab:impact-local-kd}
\footnotesize
\begin{tabular}{@{}llllll@{}}
\toprule
\multirow{2}{*}{Methods} & \multirow{2}{*}{KD} & \multicolumn{4}{l}{Accuracy}      \\ \cmidrule(l){3-6} 
&                     
& Exit 1 & Exit 2 & Exit 3 & Exit 4 \\ \midrule
\multirow{2}{*}{FedPer-EE \cite{arivazhagan2019federated}}  
& \ding{55}     
& $36.29_{\pm 5.94}$
& $37.92_{\pm 6.36}$
& $41.45_{\pm 6.02}$
& $41.88_{\pm 5.79}$    
\\
& \ding{51}   
& $36.94_{\pm 5.81}$  
& $38.79_{\pm 5.62}$
& $39.15_{\pm 6.04}$
& $38.33_{\pm 5.94}$
\\ \midrule
\multirow{2}{*}{FedRep-EE \cite{collins2021exploiting}}  
& \ding{55}      
& $34.42_{\pm 6.44}$
& $35.65_{\pm 6.16}$
& $38.46_{\pm 6.00}$
& $38.21_{\pm 5.95}$
\\
& \ding{51}      
& $34.96_{\pm 6.30}$
& $36.14_{\pm 6.39}$    
& $39.07_{\pm 6.27}$
& $38.27_{\pm 5.93}$   
\\ \midrule
\multirow{2}{*}{FedBABU-EE \cite{oh2022fedbabu}} 
& \ding{55}   
& $26.99_{\pm 7.29}$
& $36.74_{\pm 9.97}$
& $37.80_{\pm 10.66}$
& $37.32_{\pm 10.71}$
\\
& \ding{51}   
& $31.02_{\pm 7.11}$
& $32.56_{\pm 8.17}$
& $32.37_{\pm 8.71}$
& $30.82_{\pm 8.30}$
\\ \midrule
\multirow{2}{*}{FedRoD-EE \cite{chen2022bridging}}  
& \ding{55}     
& $31.90_{\pm 7.30}$
& $42.08_{\pm 7.46}$
& $48.27_{\pm 8.23}$
& $48.11_{\pm 7.15}$
\\
& \ding{51}   
& $31.68_{\pm 7.70}$
& $41.80_{\pm 7.96}$
& $47.62_{\pm 8.73}$ 
& $46.82_{\pm 6.85}$
\\ \midrule
\multirow{2}{*}{FedPAC-EE \cite{xu2023personalized}}  
& \ding{55}                  
& $38.47_{\pm 6.18}$    
& $46.54_{\pm 6.15}$   
& $52.13_{\pm 5.92}$   
& $52.67_{\pm 6.16}$   
\\
& \ding{51}                  
& $38.49_{\pm 6.34}$   
& $46.10_{\pm 6.40}$    
& $51.65_{\pm 6.25}$
& $52.10_{\pm 6.35}$         
\\ \bottomrule
\end{tabular}
\end{table}

\tabref{tab:impact-local-kd} summarizes the per-exit accuracy with and without local KD on CIFAR-100 under Dir(0.3).
With local KD, the accuracy of shallow exits (\eg Exit-1) is slightly improved.
Yet the accuracy of deeper exits degrade.
It implies that with severe data heterogeneity, local KD cannot effectively train personalized EENs.
The result aligns with our motivation of cross-client KD.

\subsubsection{Computational Cost of Teacher Matching}
\label{subsubseec:comp-teacher-matching}
We measured the teacher matching latency under different sampling scales (see \tabref{tab:teacher-matching-latency}).

\begin{table}[]
\caption{Computational time of teacher matching under different numbers of sampled clients.}
\label{tab:teacher-matching-latency}
\centering
\footnotesize
\begin{tabular}{@{}llllll@{}}
\toprule
Sampled Clients & 10 & 20 & 50 & 100 & 200 \\
\midrule
Time (s) & 0.02 & 0.02 & 0.05 & 0.48 & 1.43 \\
\bottomrule
\end{tabular}
\end{table}

Empirically, teacher matching in our experimental setup (\ie 100 clients with a sampling rate of $0.1$) takes 0.02 seconds.
The results show that even with 200 sampled clients (\eg 10,000 clients with a sampling rate of $0.02$), teacher matching takes only 1.43s, confirming negligible server overhead.

\subsubsection{Visualization of Cross-client KD Weight}
\begin{figure}[ht]
  \centering
  \subfloat[Weight under Dir(0.3)]{
    \includegraphics[width=0.2\textwidth]{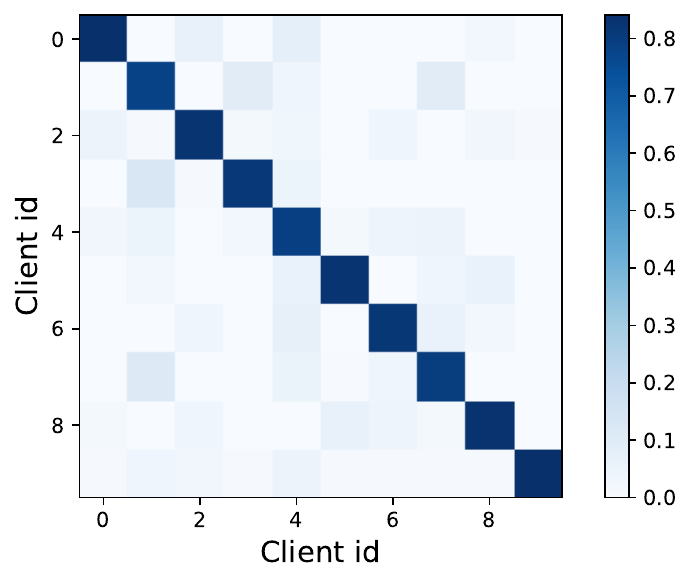}
    \label{fig:vis-100}
  }
  \subfloat[Weight under Dir(0.1)]{
    \includegraphics[width=0.2\textwidth]{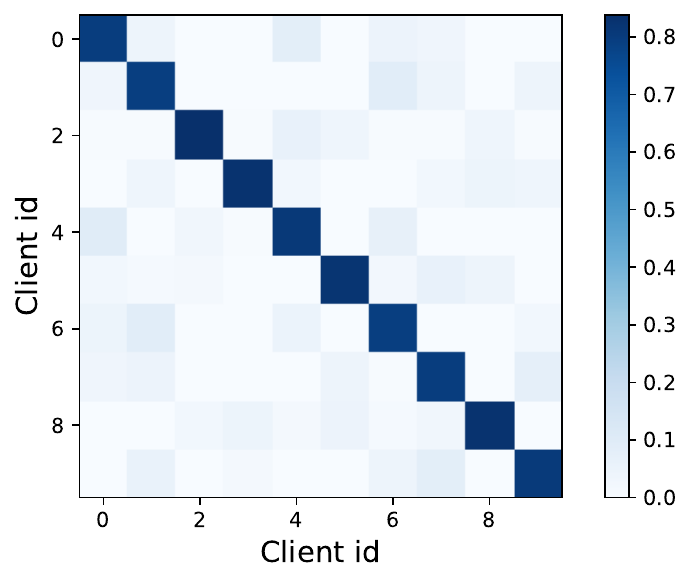}
    \label{fig:vis-100-a1}
  }
  \caption{Visualization of cross-client KD weight.}
  \label{fig:vis}
\end{figure}

We visualize the cross-client KD weight on CIFAR-100 dataset under both Dir(0.3) and Dir(0.1) in \figref{fig:vis}.
For all clients, the weight assigned to the client itself is approximately $0.8$.
This indicates that, despite cross-client KD improves the performance as demonstrated in \secref{subsec:ablation}, it still primarily depends on the client's own teacher exit.

\subsection{Ablation Studies}
\label{subsec:ablation}

\subsubsection{Contributions of Individual Components}
\tabref{tab:ablation} compares four variants of \sysname on CIFAR-100 and TinyImageNet under Dir(0.3).
Naively adding local KD does not improves accuracy, which aligns with \secref{subsubsec:necessity}.
With cross-client KD (\ie cross-client teacher matching in \secref{subsec:teacher-matching}), the accuracy of all exits increases by $5.77$-$10.03\%$ on CIFAR-100 and $4.24$-$8.48\%$ on TinyImageNet.
Adding conflict-aware student selection in \secref{subsec:scheduling} further improves the performance of all exits by $1.17$-$5.21\%$ and $1.17$-$3.09\%$.


\begin{table}[t]
\caption{Contributions of individual components. 
KD: naive local 
\underline{k}nowledge 
\underline{d}istillation; 
TM: cross-client 
\underline{t}eacher 
\underline{m}atching; 
SS: conflict-aware 
\underline{s}tudent 
\underline{s}election.}
\footnotesize
\label{tab:ablation}
\begin{tabular}
{@{}llllllll@{}}
\toprule
\multirow{2}{*}{Dataset}      
& \multicolumn{3}{l}{Options} 
& \multicolumn{4}{l}{Accuracy}                 
\\ 
\cmidrule(l){2-8} 
& KD    & TM    & SS   
& Exit 1 & Exit 2 & Exit 3 & Exit 4 
\\ \midrule
\multirow{4}{*}{CIFAR-100}    
& \ding{55}   
& \ding{55}   
& \ding{55}  
& $36.29_{\pm 5.94}$
& $37.92_{\pm 6.36}$
& $41.45_{\pm 6.02}$
& $41.88_{\pm 5.79}$
\\
& \ding{51}         
& \ding{55}   
& \ding{55}  
& $36.45_{\pm 5.65}$
& $38.10_{\pm 5.67}$
& $38.31_{\pm 5.63}$
& $37.51_{\pm 5.66}$ 
\\
& \ding{51}     
& \ding{51}    
& \ding{55}    
& $42.06_{\pm 6.14}$
& $45.64_{\pm 6.26}$
& $48.21_{\pm 5.54}$
& $51.91_{\pm 5.18}$
\\
& \ding{51}      
& \ding{51}    
& \ding{51}   
& $45.45_{\pm 6.06}$
& $49.69_{\pm 5.96}$
& $53.42_{\pm 5.96}$
& $53.08_{\pm 6.26}$
\\ \midrule
\multirow{4}{*}{
\begin{tabular}[c]{@{}l@{}}Tiny-\\ ImageNet\end{tabular}
} 
& \ding{55}   
& \ding{55}   
& \ding{55}  
& $26.20_{\pm 12.34}$
& $26.41_{\pm 12.08}$
& $27.80_{\pm 11.98}$
& $29.80_{\pm 11.65}$           
\\
& \ding{51}         
& \ding{55}   
& \ding{55}   
& $25.27_{\pm 12.12}$
& $25.75_{\pm 11.97}$
& $26.35_{\pm 12.09}$ 
& $26.52_{\pm 12.22}$
\\
& \ding{51}     
& \ding{51}    
& \ding{55}  
& $30.44_{\pm 11.95}$
& $32.16_{\pm 11.58}$
& $34.85_{\pm 10.98}$ 
& $38.28_{\pm 10.45}$
\\
& \ding{51}      
& \ding{51}    
& \ding{51}     
& $31.61_{\pm 11.71}$
& $34.39_{\pm 11.42}$
& $37.72_{\pm 10.73}$
& $41.37_{\pm 10.09}$
\\ 
\bottomrule
\end{tabular}
\end{table}

\subsubsection{Round-to-Accuracy}
\begin{figure}[t]
  \centering
  \subfloat[CIFAR-100]{
    \includegraphics[width=0.23\textwidth]{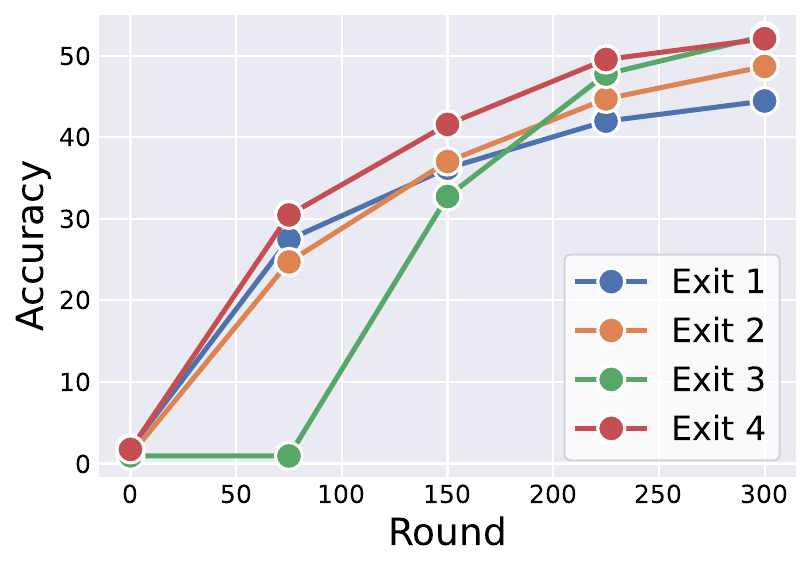}
    \label{fig:round2acc-cifar100}
  }
  \subfloat[TinyImageNet]{
    \includegraphics[width=0.23\textwidth]{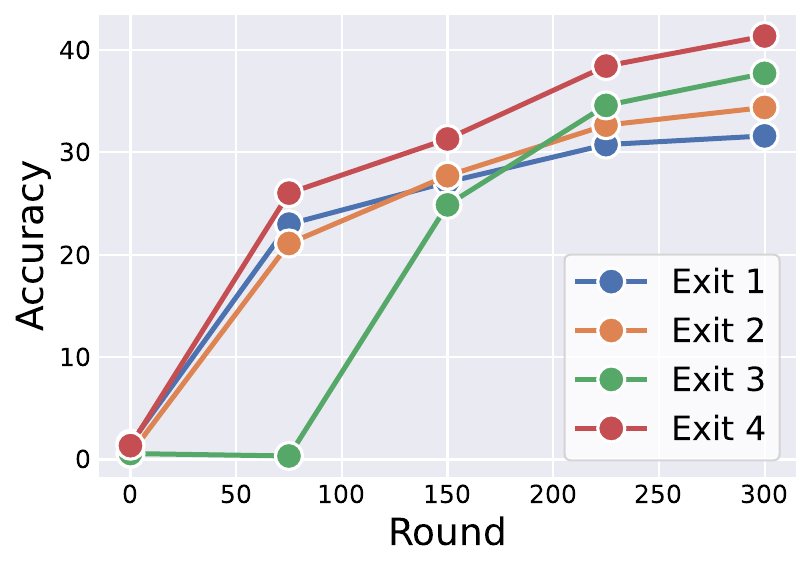}
    \label{fig:round2acc-tiny}
  }
  \caption{Round-Accuracy curves.}
  \label{fig:impact-round2acc}
\end{figure}

\figref{fig:impact-round2acc} plots the round-to-accuracy curves of all exits.
The accuracy of each exit increases asynchronously.
The teacher exit \ie Exit 4, consistently yields the highest accuracy.
Initially, Exit 1 outperforms Exit 2 and Exit 3 because the latter two have not been selected yet.
As the training progresses, Exit 3 gradually surpasses Exit 1 and 2.
It implies that our exit selection mechanism does not compromise the performance of deeper exits, even though they are only involved in the later stage of training.

\subsubsection{Impact of Selection Rate $R(t)$}
\label{subsubsec:impact-selection-rate}
\begin{figure}[t]
  \centering
  \subfloat[CIFAR-100]{
    \includegraphics[width=0.23\textwidth]{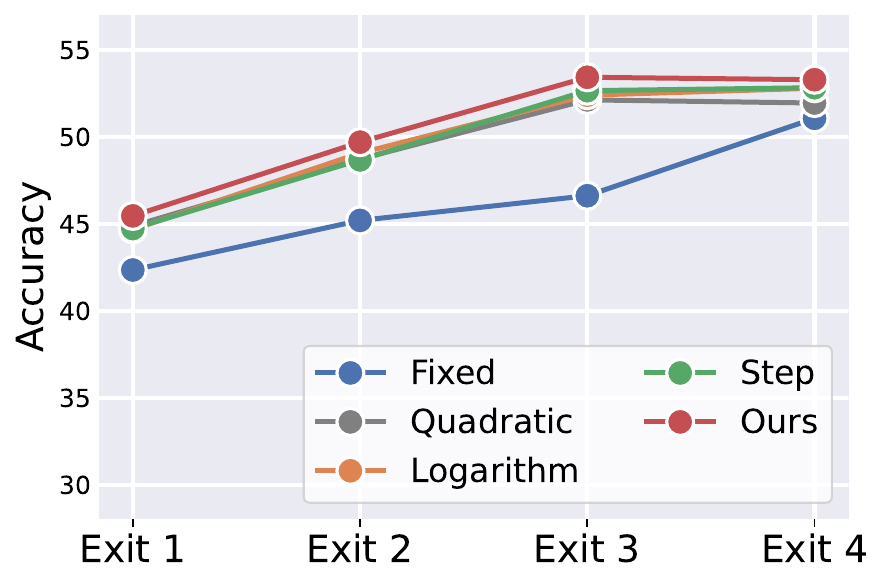}
    \label{fig:rt-cifar100}
  }
  \subfloat[TinyImageNet]{
    \includegraphics[width=0.23\textwidth]{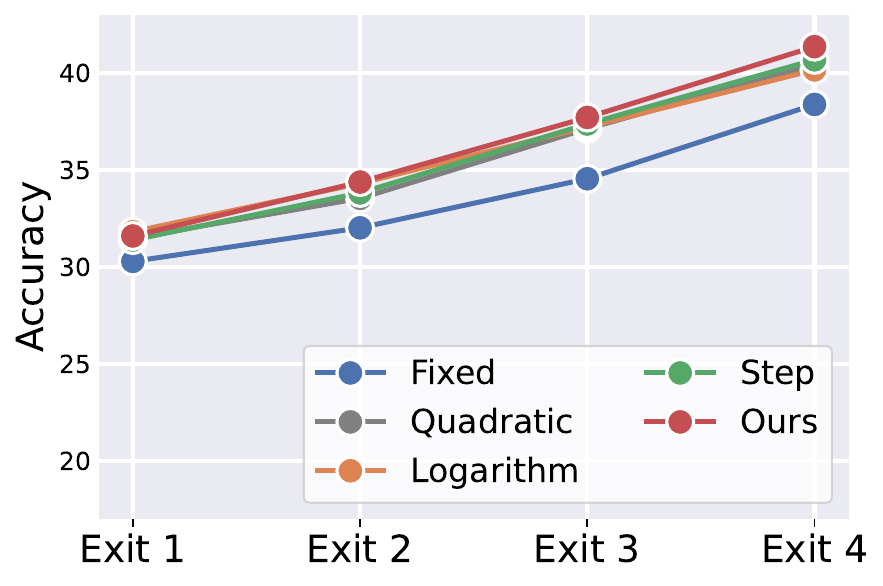}
    \label{fig:rt-tinyimagenet}
  }
  \caption{Impact of selection rate $R(t)$.}
  \label{fig:impact-rt}
\end{figure}

This experiment compares different options for the selection rate $R(t)$.
Given $T$ rounds, \sysname increases $R(t)$ linearly, \ie $R(t)=\min(2t/T,1)$ (see \secref{subsec:scheduling}).
We explore four alternatives:
(1) \textit{Fixed}: $R(t)=1$.
(2) \textit{Quadratic}: $R(t)=\min((2t/T)^2,1)$.
(3) \textit{Logarithm}: $R(t)=\min(\log (2(e-1)t/T+1),1)$.
(4) \textit{Step}: Every $T/20$ rounds, $R(t)$ increases by 0.1 until it reaches $1$.
\figref{fig:impact-rt} presents the per-exit accuracies of EENs trained with different $R(t)$.
All strategies except \textit{fixed} (\ie w/o exit selection) achieve comparable performance.
The worst among the four strategies still outperforms \textit{fixed} by $2.33\%$, $3.48\%$, $5.51\%$ and $0.92\%$ for CIFAR-100, and $1.10\%$, $1.52\%$, $2.58\%$ and $1.78\%$ for TinyImageNet.
\zimu{We pick the linear strategy for its simplicity.}

\subsubsection{Impact of Hyperparameters}
\begin{figure}[t]
  \centering
  \subfloat[CIFAR-100 with $\mu$]{
    \includegraphics[width=0.23\textwidth]{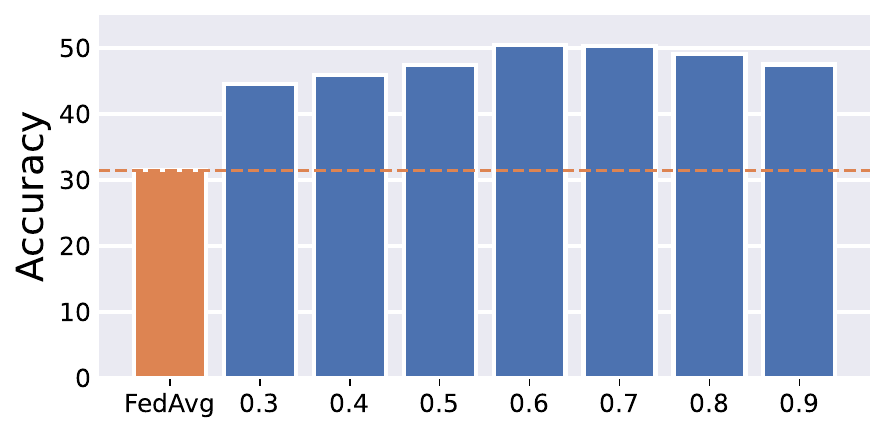}
    \label{fig:hyper-cifar100-alpha}
  }
  \subfloat[CIFAR-100 with $\lambda$]{
    \includegraphics[width=0.23\textwidth]{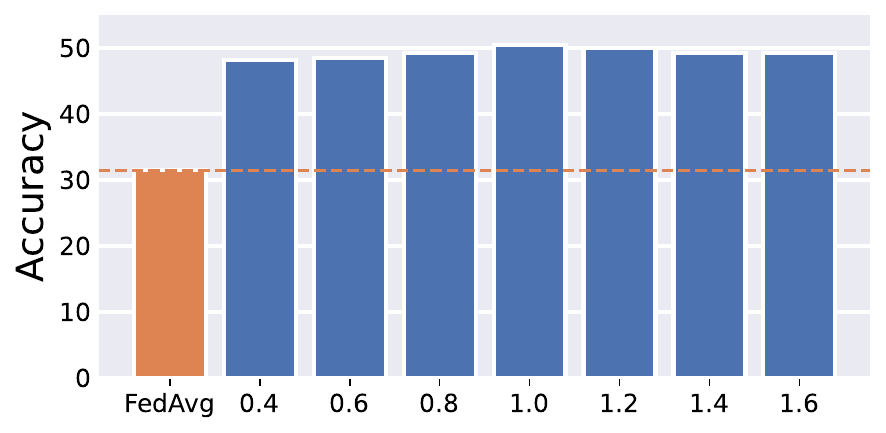}
    \label{fig:hyper-cifar100-lam}
  }
  
  \subfloat[TinyImageNet with $\mu$]{
    \includegraphics[width=0.23\textwidth]{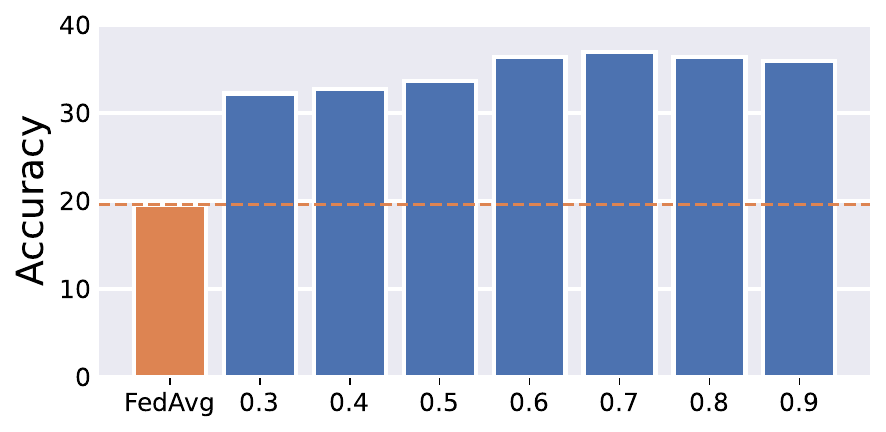}
    \label{fig:hyper-tiny-alpha}
  }
  \subfloat[TinyImageNet with $\lambda$]{
    \includegraphics[width=0.23\textwidth]{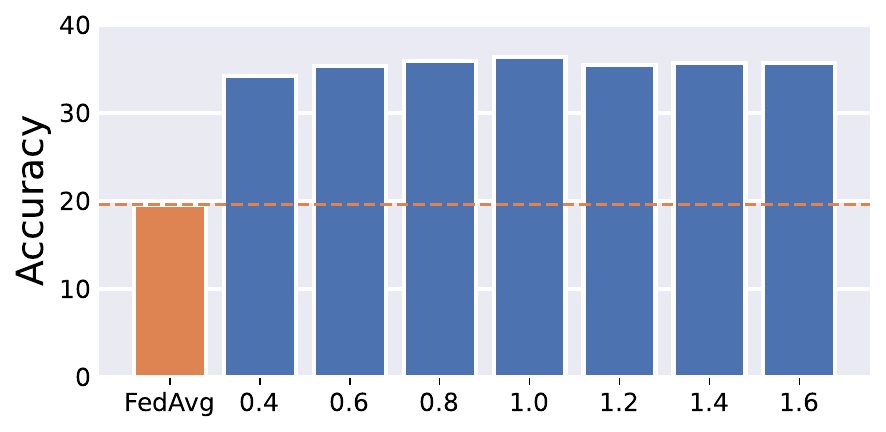}
    \label{fig:hyper-tiny-lam}
  }
  \caption{Impact of hyperparameters.}
  \label{fig:impact-hyper}
  \vspace{-3mm}
\end{figure}

\figref{fig:impact-hyper} shows the hyperparameter sensitivity of \sysname.
We test two key hyperparameters: $\mu$ that regularizes the cross-client KD weight $\vk$ in \equref{equ:solve-k}, and $\lambda$ for KD in \equref{equ:overall-obj}.
We vary $\mu$ from $0.3$ to $0.9$, and the optimal performance is attained when $\mu \in [0.6,0.8]$.
A large $\mu$ uniforms the cross-client KD weight.
Teacher exits from irrelevant clients may have a greater influence on local training.
Conversely, a small $\mu$ forces cross-client KD into local KD.
Regarding $\lambda$, \sysname demonstrates robust performance across a range from $0.4$ to $1.6$.

\subsubsection{Impact of Exit Policy}
\label{subsubsec:impact-policy}
\begin{figure}[t]
  \centering
  \subfloat[Accuracy vs. MAC]{
    \includegraphics[width=0.2\textwidth]{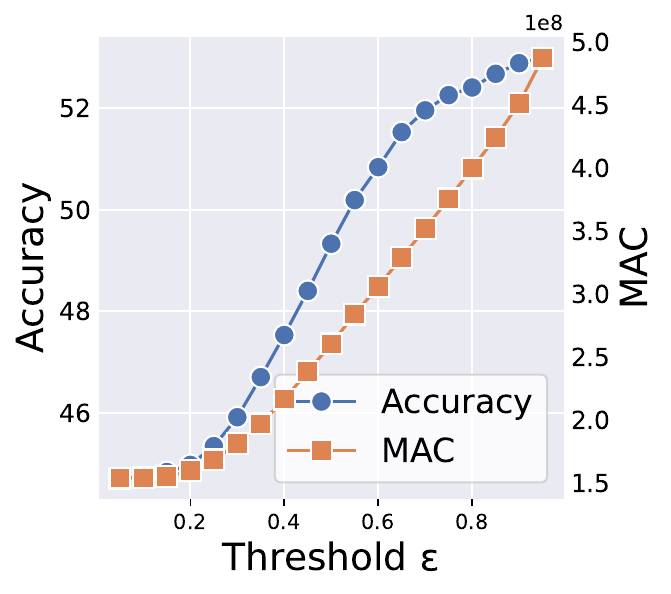}
    \label{fig:policy-dual}
  }
  \subfloat[Round vs. MAC]{
    \includegraphics[width=0.27\textwidth]{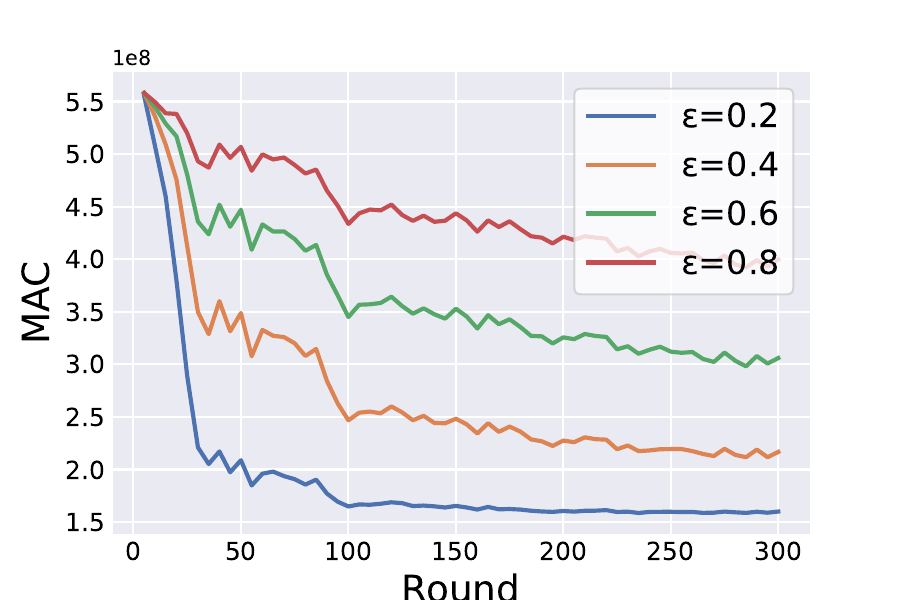}
    \label{fig:policy-timeacc}
  }
  \caption{Impact of threshold $\epsilon$ in exit policy.}
  \label{fig:impact-policy}
\end{figure}

We evaluate \sysname under different exit thresholds $\epsilon$ on CIFAR-100 with Dir(0.3).
\figref{fig:policy-dual} shows that both accuracy and inference cost increase with the threshold $\epsilon$. 
This is because the model is likely to terminate at a deeper exit given high threshold.
We find $\epsilon = 0.6$ better balance between inference accuracy and efficiency.
Compared to full-model inference, there is an accuracy drop of $2.01\%$ for \sysname, yet the number of MACs is reduced by $37.15\%$.
\figref{fig:policy-timeacc} shows that the inference cost drops during training given different thresholds.
This is because shallow exits become increasingly more confident in their outputs.

\subsection{Case Study}
\begin{figure}
    \centering
    \subfloat[Deployment]
    {   \includegraphics[width=0.23\linewidth]{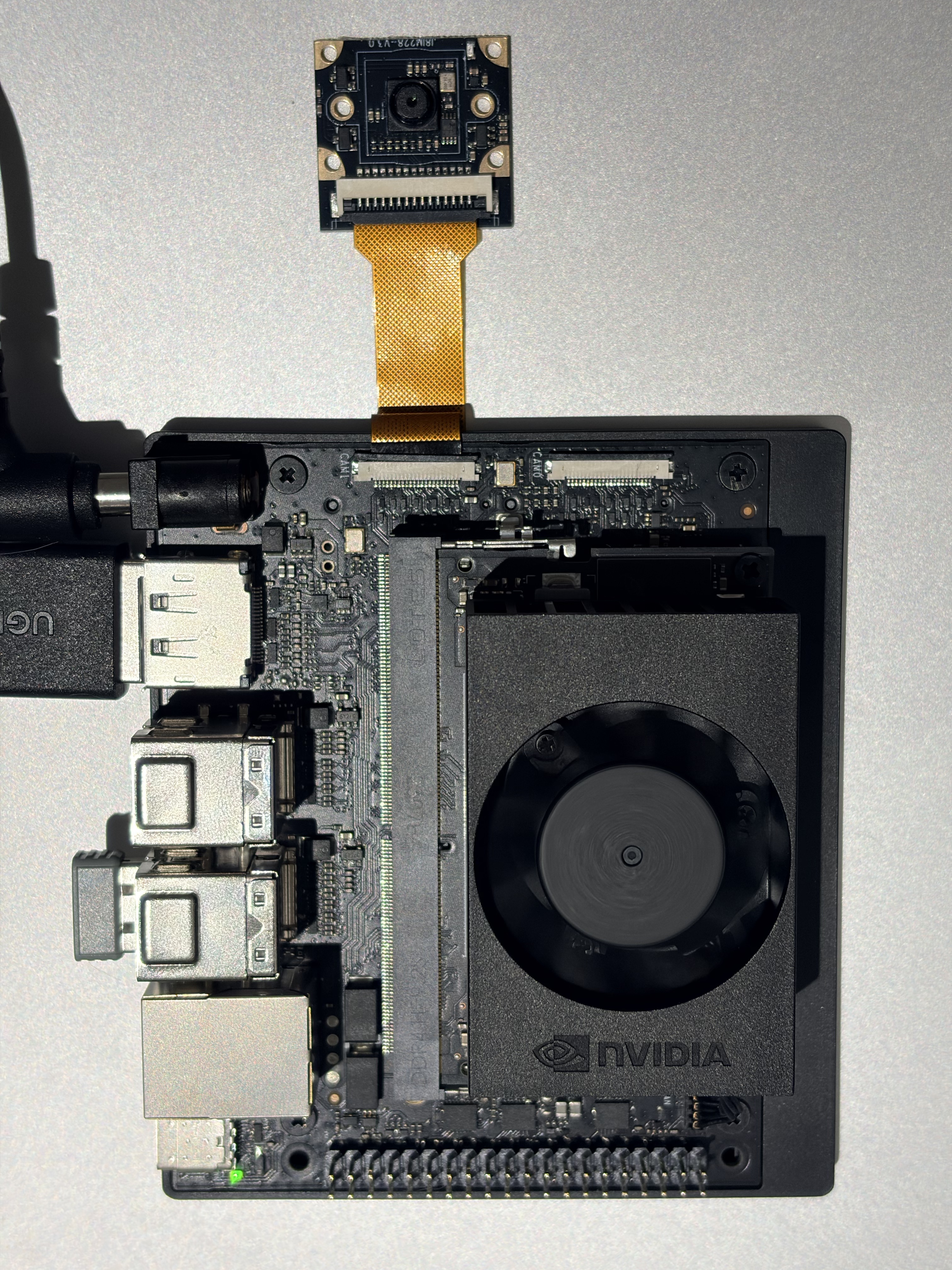}
    \label{fig:jetson}
    }
    \subfloat[CIFAR-100]
    {   \includegraphics[width=0.38\linewidth]{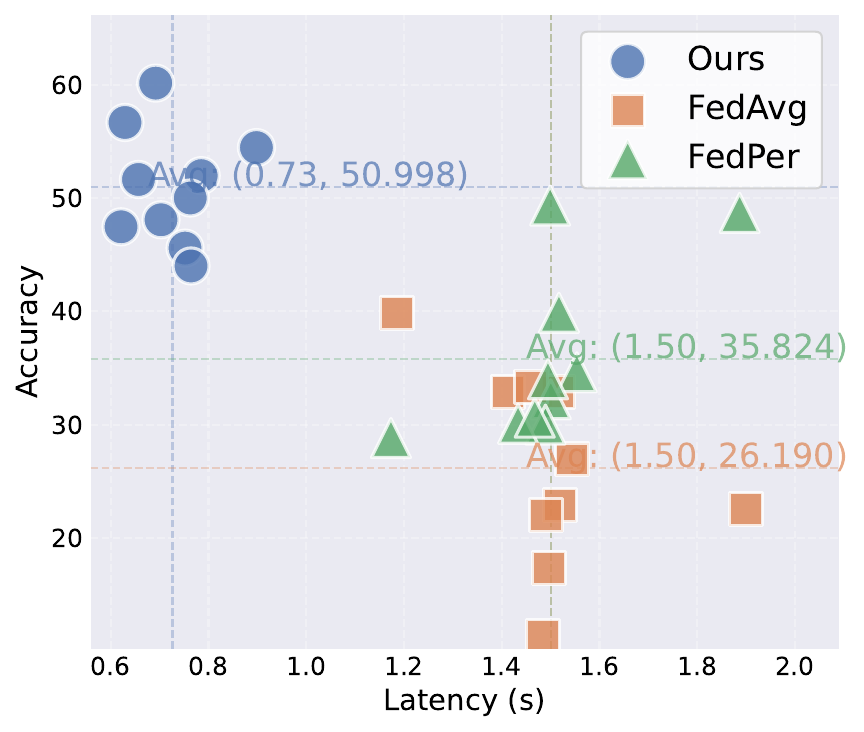}
    \label{fig:case-easy}
    }
    \subfloat[CIFAR-100 with Noise]
    {   \includegraphics[width=0.38\linewidth]{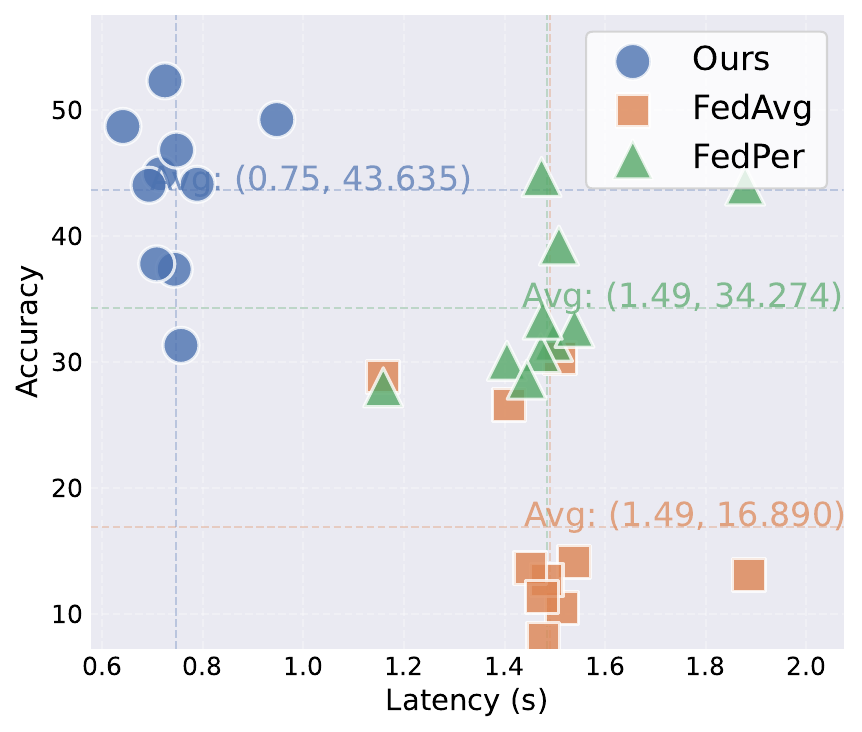}
    \label{fig:case-hard}
    }
    
    \caption{Case study on edge devices: (a) deployment on Jetson Nano; accuracy-latency tradeoff on (b) CIFAR-100 and (c) CIFAR-100 with noise.}
    \label{fig:case}
\end{figure}

\figref{fig:case} presents a case study by deploying the trained models of \sysname, FedAvg \cite{mcmahan2017communication} and FedPer \cite{arivazhagan2019federated} on an edge device (Jetson-Nano with 8GB Memory, see \figref{fig:jetson}).
For clear illustration, we only report the results of the first 10 clients out of 100 clients in training.
We evaluate \textit{(i) averaged accuracy} and \textit{(ii) averaged inference latency} of each client on the corresponding testset.
We choose ResNet-18 as model, and CIFAR-100 as dataset (see \figref{fig:case-easy}). 
We also add Gaussian noise with $\sigma=0.1$ to the test images to simulate noisy environments (see \figref{fig:case-hard}).
The exit confidence threshold is set to $0.5$.
Compared with the baselines, \sysname achieves higher accuracy with lower inference latency on most clients.
Specifically, \sysname yields an accuracy gain of $15.17$-$24.81\%$, $9.36$-$26.66\%$ and reduces inference latency by $2.05\times$, $1.99\times$ in CIFAR-100, and CIFAR-100 with noise.
The latency is slightly higher in the noisy setting because the model often fails to produce confident outputs at early exits.

\section{Conclusion}

In this paper, we propose \sysname, a Conflict-Aware Cross-Client Federated Exit Distillation framework that extends personalized federated learning to early-exit networks. 
By jointly addressing client-wise heterogeneity and depth-wise interference, \sysname harmonizes exit training in a unified framework. 
Its progressive, depth-prioritized coordination strategy mitigates conflicts between shallow and deep exits and allows effective knowledge transfer across clients, while the client-decoupled formulation avoids excessive communication overhead. 
Extensive experiments on four benchmark datasets show that \sysname outperforms state-of-the-art PFL and GFL-EE baselines in accuracy and reduces inference costs by 30.79\%–46.86\%. 
We envision \sysname as a first step towards personalized and dynamic models from decentralized, heterogeneous IoT data.
In the future, we plan to explore optimizations for client resource heterogeneity and expand \sysname to multimodal scenarios.


\section*{Acknowledgments}
This research is partially supported by 
National Natural Science Foundation of China (NSFC) (No. 62572412), 
the RGC of Hong Kong SAR, China (Project No. CityU 11206425), 
and City University of Hong Kong Shenzhen Research Institute (CityUHKSRI).
This work is partially supported by National Natural Science Foundation of China (NSFC) (Grant Nos. 62425202, 62336003), the Beijing Natural Science Foundation (Z230001), the Fundamental Research Funds for the Central Universities No.JKF-2026007844235, and the State Key Laboratory of Complex \& Critical Software Environment (SKLCCSE).
Zimu Zhou and Yongxin Tong are the corresponding authors.

\appendix




\section{Appendix}

\subsection{Proof of \propref{prop:equv}}
\label{proof:equv}
\begin{proof}
    Taking the definition of KL divergence into \equref{equ:cc-kd}:
    \begin{equation}
    \label{equ:theta-optim-flatten}
    \begin{aligned}
        \theta^* 
        &= \textstyle \arg \min_{\theta} \sumn k_i \theta_{im} (x) \log \frac{\theta_{im} (x)}{\theta(x)}  \\
        &= \textstyle \arg \min_{\theta} - \sumn k_{i} \theta_{im} (x) \log \theta(x) \\
        &= \textstyle \arg \min_{\theta} -\log\theta(x) \sumn k_i \theta_{im} (x)
    \end{aligned}
    \end{equation}

    The model $\theta_{im}$ can be further expressed as $\theta_{im}=(\phi,h_{im})$.
    The global backbone $\phi$ is shared by all clients.
    The linear classifier $h_{im}$ is kept locally.
    Let the model with the aggregated classifier be $\bar{\theta}=(\phi, \bar{h})$.
    The averaged classifier matrix is $\bar{h} = \sumn k_i h_{im}$.
    Let the output of block $m$ of backbone $\phi$ be $\phi(x)$, we have:
    \begin{equation}
    \label{equ:expansionhim}
    \begin{aligned}
        \bar{\theta}(x) \textstyle
        = \bar{h}(\phi(x)) 
        = \phi(x)^\top \bar{h}  
        = \phi(x)^\top \sumn k_{i} h_{im}  
        = \sumn k_{i} \phi(x)^\top h_{im}
    \end{aligned}
    \end{equation}
    Since the backbone $\phi$ is shared across all clients, the output of exit $\theta_{im}$ can be represented by $\theta_{im}(x) = h_{im} (\phi(x)) = \phi(x)^\top h_{im}$.
    Taking it into \equref{equ:expansionhim}, there is $\textstyle \bar{\theta}(x) = \sumn k_i \theta_{im}(x)$.
    Taking it into \equref{equ:theta-optim-flatten}, we have:
    \begin{equation}
    \label{equ:kdijnew}
    \begin{aligned}
        \theta^* &
        = \textstyle \arg \min_{\theta} - \bar{\theta}(x) \log \theta(x)  \\
        &= \textstyle \arg \min_{\theta} \bar{\theta}(x) (\log\bar{\theta}(x) - \log \theta(x)) = \textstyle \arg \min_{\theta} D_{KL}(\bar{\theta} || \theta)
    \end{aligned}
    \end{equation}

    Combining \equref{equ:cc-kd} and \equref{equ:kdijnew}, we complete the proof.
\end{proof}

\subsection{Convergence Analysis of \sysname}
\label{proof:convergence}

\newcommand{\Fat}{F(\phi^t, h^t)}
\newcommand{\Fi}{F_i(\phi, h_{i})}
\newcommand{\Fit}{F_i(\phi^t, h_{i}^t)}
\newcommand{\nah}{\nabla_h}
\newcommand{\nap}{\nabla_\phi}
\newcommand{\tip}{\tilde{\phi}}
\newcommand{\tih}{\tilde{h}}
\newcommand{\sumk}{\sum_{e=1}^E}
\newcommand{\Fitt}{F_i(\tip_{i}^{t,e}, \tih_{i}^{t,e})}
\newcommand{\met}{\mathbb{E}}
\newcommand{\divp}{\tilde{\phi}_i^{t,e} - \phi^t}
\newcommand{\divh}{\tilde{h}_i^{t,e} - h_i^t}
\newcommand{\etat}{\eta}

\begin{theorem}
\label{thm:convergence}
    (Convergence of \sysname). Under \asmref{asm:smooth}--\asmref{asm:bound}, \sysname satisfies:
\begin{equation}
\begin{aligned}
    \textstyle \frac{1}{T}\sum_{t=1}^{T}
    ( \met \left\| \nap \Fat \right\|^2
    + s \met  \left\| \nah \Fit \right \| ^2
    )
     \textstyle \le \frac{2\Delta F_0}{E\etat T} + \eta \Phi_1 + \eta^2 \Phi_2 + O(\delta_i)
\end{aligned}
\end{equation}
with $\Phi_1 = \textstyle \frac{L_\phi (1+\chi^2) E^2 (G_\phi^2+\sigma_\phi^2)}{2} + \textstyle\frac{2sL_h(1+\chi^2)}{n} ( 4(E-1)^2 (G_h^2+\sigma_h^2)   + E^2 (G_h^2+\sigma_h^2))$ and $\Phi_2 = \textstyle \frac{E^3}{3} (L_\phi^2  (G_\phi^2+\sigma_\phi^2) + \chi^2 L_h L_\phi  (G_h^2+\sigma_h^2)) \textstyle + sE^3 (\chi^2 L_\phi L_h  (G_\phi^2+\sigma_\phi^2) + L_h^2  (G_h^2+\sigma_h^2) )$.
\end{theorem}

\fakeparagraph{Assumptions}
Let $\phi$ be the backbone and $h_i=\{h_{ij}|j\in\mathbb{S}_i\}$ the exits of client $i$.
We adopt the standard assumptions in FL convergence analysis~\cite{li2020convergence,pillutla2022federated}, adapted to our two-block (backbone--exit) objective:

\begin{assumption}
\label{asm:smooth}
    (Smoothness). $F_i$ is block-wise $L$-smooth: $\phi\mapsto\nap\Fi$ is $L_\phi$-smooth, $h_i\mapsto\nah\Fi$ is $L_h$-smooth, and the cross terms satisfy $\max\{L_{\phi h},L_{h\phi}\}\le\chi\sqrt{L_h L_\phi}$ for some $\chi>0$.
\end{assumption}

\begin{assumption}
\label{asm:unbiased}
    (Unbiased gradient). $\met_{\xi_i\sim D_i}\nabla F_i(\phi,h_i;\xi_i)=\nabla F_i(\phi,h_i)$ for both $\nap$ and $\nah$.
\end{assumption}

\begin{assumption}
\label{asm:bound}
    (Bounded variance and gradient). For both $\nap$ and $\nah$, $\met\|\nabla F_i(\phi,h_i;\xi_i)-\nabla F_i(\phi,h_i)\|^2\le\sigma_\bullet^2$ and $\met\|\nabla F_i(\phi,h_i)\|^2\le G_\bullet^2$, with $\bullet\in\{\phi,h\}$.
\end{assumption}

\fakeparagraph{Proof Outline}
The overall objective is $F(\phi^t, h^t) = \textstyle \sumn p_i F_i(\phi^t, h_i^t)$.
Following Lemma~20 of~\cite{pillutla2022federated}, we have the per-round descent:
\begin{equation}
\label{equ:decompose}
\begin{aligned}
    &F(\phi^{t+1}, h^{t+1}) - F(\phi^t, h^t) \\
    \le &
    \underbrace{\langle \nap \Fat, \phi^{t+1}-\phi^t \rangle }_{A_1}
    + \underbrace{\tfrac{1}{n} \textstyle \sumn \left\langle \nah \Fit, h_i^{t+1} - h_i^t \right\rangle}_{B_1} \\
    & + \underbrace{\tfrac{L_\phi (1+\chi^2)}{2} \left\| \phi^{t+1} - \phi^t\right\|^2}_{A_2}
    + \underbrace{\tfrac{1}{n} \textstyle \sumn \frac{L_h(1+\chi^2)}{2} \| h_i^{t+1} - h_i^{t} \|^2}_{B_2}
\end{aligned}
\end{equation}
$A_1, A_2$ concern the shared backbone and follow standard FedAvg arguments~\cite{li2020convergence,pillutla2022federated}. The novelty lies in $B_1, B_2$, which involve \sysname's cross-client KD update of teacher exits with weights $k_j^i$.

\begin{lemma}
    \label{lemma:div}
    (Local drift, standard~\cite{pillutla2022federated}). For backbone $\tilde{\phi}_i^{t,e}$ and exits $\tilde{h}_i^{t,e}$ at local epoch $e$:
    \begin{equation}
        \met \| \divp \|^2 \le  (e-1)^2 \etat^2 (G_\phi^2+\sigma_\phi^2), \quad \met  \| \divh \|^2  \le (e-1)^2\etat^2 (G_h^2+\sigma_h^2)
    \end{equation}
\end{lemma}
The proof follows by writing each drift as a telescoping sum of one-step updates, applying Jensen's inequality, and using \asmref{asm:unbiased}--\asmref{asm:bound}; we omit it (cf.~\cite{pillutla2022federated}).
\begin{equation}
\label{equ:phi-bound}
    \met \| \textstyle \sumk \nap F_i(\tip_i^{t,e}, h_i;\xi_i)\|^2 \le E^2(G_\phi^2+\sigma_\phi^2)
\end{equation}
follows as a corollary by same argument for cumulative gradient.

\begin{lemma}
\label{lemma:a1}
(Bounding $A_1$).
$\met [A_1] \le
    \textstyle -\frac{\eta E}{2} \met\| \nap \Fat \|^2
    + \textstyle \frac{(E\etat)^3}{3}  (L_\phi^2  (G_\phi^2+\sigma_\phi^2) + \chi^2 L_h L_\phi  (G_h^2+\sigma_h^2) )$.
\end{lemma}
\begin{lemma}
    \label{lemma:a2}
    (Bounding $A_2$). $\met [A_2] \le \textstyle \frac{L_\phi (1+\chi^2) \etat^2 E^2 (G_\phi^2+\sigma_\phi^2)}{2}$.
\end{lemma}
\lemmaref{lemma:a1} follows by introducing a positive term, applying $-\langle a,b\rangle\le\frac{1}{2}\|a\|^2+\frac{1}{2}\|b\|^2$, smoothness (\asmref{asm:smooth}), and \lemmaref{lemma:div}---a derivation identical to standard FedAvg analysis on a single block~\cite{pillutla2022federated}. \lemmaref{lemma:a2} follows directly from \equref{equ:phi-bound} and Jensen.

\begin{lemma}
    \label{lemma:b1}
    (Bounding $B_1$).
    $\met [B_1] \le \textstyle - \frac{s\etat E}{2} \met \| \nah \Fit \|^2 + \eta G_h \delta_i +  s(E\etat)^3 (\chi^2 L_\phi L_h (G_\phi^2+\sigma_\phi^2) + L_h^2 (G_h^2+\sigma_h^2))$.
\end{lemma}
\begin{proof}
    Only the $s$-fraction of clients in $\mathbb{C}$ update exits, so $h_i^{t+1}-h_i^t=0$ for the rest.
    Split $B_1=\frac{1}{n}\sum_{i\in\mathbb{C}}(B_{11}+B_{12})$ where
    $B_{11}=\langle \nah\Fit, h_{im}^{t+1}-h_{im}^{t,E}\rangle$ captures the cross-client KD aggregation (novel) and
    $B_{12}=\langle \nah\Fit, h_i^{t,E}-h_i^t\rangle$ captures the local drift (analogous to $A_1$).

    For $B_{11}$, substituting $h_{im}^{t+1}=\sum_{j\in\mathbb{C}} k_j^i h_{jm}^{t,E}$ produces three inner products over $j\neq i$:
    \begin{equation}
        \begin{aligned}
            \met[B_{11}] =& \textstyle \sum_{j\in\mathbb{C},j\neq i} k_j^i \big[
            \langle\nah\Fit, h_{jm}^{t,E}-h_{jm}^t\rangle \\
            &+ \langle\nah\Fit, h_{jm}^t-h_{im}^t\rangle 
            + \langle\nah\Fit, h_{im}^t-h_{im}^{t,E}\rangle \big].
        \end{aligned}
    \end{equation}
    The first and third terms are bounded by $\frac{(E\etat)^3}{3}(\chi^2 L_\phi L_h(G_\phi^2+\sigma_\phi^2)+L_h^2(G_h^2+\sigma_h^2))$ each, via $-\langle a,b\rangle\le\frac{1}{2}\|a-b\|^2$, smoothness, and \lemmaref{lemma:div} (note $\nah F_{im}$ is a special case of $\nah F_i$).
    The second term is the cross-client heterogeneity contribution. Defining $\delta_i := \sum_{j\in\mathbb{C},j\neq i} k_j^i \| h_{jm}^t-h_{im}^t \|^2/\eta$ as the system data heterogeneity, Cauchy--Schwarz and \asmref{asm:bound} give $\le \eta G_h\delta_i$.
    For $B_{12}$, the same argument as \lemmaref{lemma:a1} yields $-\frac{\etat E}{2}\met\|\nah\Fit\|^2+\frac{(E\etat)^3}{3}(\chi^2 L_\phi L_h(G_\phi^2+\sigma_\phi^2)+L_h^2(G_h^2+\sigma_h^2))$.
    Summing $B_{11}$ and $B_{12}$ completes the proof.
\end{proof}

\begin{lemma}
\label{lemma:b2}
    (Bounding $B_2$).
    $\met[B_2] \le \textstyle \frac{2sL_h(1+\chi^2)}{n}( 2\eta \delta_i + 4(E-1)^2\eta^2 (G_h^2+\sigma_h^2) + \etat^2 E^2 (G_h^2+\sigma_h^2))$.
\end{lemma}
\begin{proof}
    Apply $\|h_i^{t+1}-h_i^t\|^2 \le 2\|h_{im}^{t+1}-h_{im}^{t,E}\|^2 + 2\|h_i^{t,E}-h_i^t\|^2$, denoting the two parts $B_{21}$ and $B_{22}$.
    For $B_{21}$, expanding $h_{im}^{t+1}=\sum_{j\in\mathbb{C}}k_j^i h_{jm}^{t,E}$ and twice applying $\|a+b\|^2\le 2\|a\|^2+2\|b\|^2$ gives
    \begin{equation}
        B_{21}\le 2 \textstyle \sum_{j\in\mathbb{C}}k_j^i (2\|h_{jm}^t-h_{im}^t\|^2 + 4\|h_{jm}^{t,E}-h_{jm}^t\|^2 + 4\|h_{im}^t-h_{im}^{t,E}\|^2),
    \end{equation}
    which by the definition of $\delta_i$ and \lemmaref{lemma:div} is at most $4\eta\delta_i+8(E-1)^2\eta^2(G_h^2+\sigma_h^2)$.
    $B_{22}$ follows the same Jensen argument as \lemmaref{lemma:a2} and is bounded by $2\etat^2 E^2(G_h^2+\sigma_h^2)$.
    Combining yields the result.
\end{proof}

Substituting \lemmaref{lemma:a1}--\lemmaref{lemma:b2} into \equref{equ:decompose} bounds $F(\phi^{t+1},h^{t+1})-F(\phi^t,h^t)$. Telescoping over $t=0,\ldots,T-1$ completes the proof.

\section*{GenAI Usage Disclosure}
The author used GenAI to revise code and polish the paper.


\bibliographystyle{ACM-Reference-Format}
\bibliography{ref}

\end{document}